\documentclass[journal]{IEEEtran}
\let\labelindent\relax
\usepackage{xcolor,soul,framed} 
\colorlet{shadecolor}{yellow}
\usepackage[pdftex]{graphicx}
\graphicspath{{../pdf/}{../jpeg/}}
\DeclareGraphicsExtensions{.pdf,.jpeg,.png}

\usepackage[cmex10]{amsmath}
\usepackage{array}
\usepackage{mdwmath}
\usepackage{mdwtab}
\usepackage{eqparbox}
\usepackage{url}
\usepackage{enumitem}
\usepackage{multirow}
\usepackage{amsmath} 
\usepackage{amssymb}
\usepackage{mathrsfs}
\usepackage{pifont}
\usepackage{bbding}
\usepackage{threeparttable}
\usepackage{booktabs}

\usepackage{graphicx}

\usepackage[caption=false,font=footnotesize,labelfont=rm,textfont=rm]{subfig}
\usepackage{stfloats}
\usepackage{cite}
\begin{document}
\bstctlcite{IEEEexample:BSTcontrol}
    \title{Hyperspectral Mamba for Hyperspectral Object Tracking}
\author{
      Long Gao\textsuperscript{a*},
      Yunhe Zhang\textsuperscript{a},
      Yan Jiang\textsuperscript{b}, 
      Weiying Xie\textsuperscript{a}, 
      Yunsong Li\textsuperscript{a},
      
\thanks{\textsuperscript{a} State Key Laboratory of Integrated Service Networks, School of Telecommunications Engineering, Xidian University, No.2, South Taibai Street, Hi-Tech Development Zone, Xi’an, China, 710071.} 
\thanks{\textsuperscript{b} The Department of Electronic and Electrical Engineering, the University of Sheffield, Sheffield, UK, S10 2TN.}
\thanks{\textsuperscript{*} Corresponding author: \emph{Long Gao}}
}

\markboth{IEEE TRANSACTIONS ON IMAGE PROCESSING
}{Roberg \MakeLowercase{\textit{et al.}}: High-Efficiency Diode and Transistor Rectifiers}




\maketitle

\begin{abstract}
Hyperspectral object tracking holds great promise due to the rich spectral information and fine-grained material distinctions in hyperspectral images, which are beneficial in challenging scenarios. While existing hyperspectral trackers have made progress by either transforming hyperspectral data into false-color images or incorporating modality fusion strategies, they often fail to capture the intrinsic spectral information, temporal dependencies, and cross-depth interactions. To address these limitations, a new hyperspectral object tracking network equipped with Mamba (HyMamba), is proposed. It unifies spectral, cross-depth, and temporal modeling through state space modules (SSMs). The core of HyMamba lies in the Spectral State Integration (SSI) module, which enables progressive refinement and propagation of spectral features with cross-depth and temporal spectral information. Embedded within each SSI, the Hyperspectral Mamba (HSM) module is introduced to learn spatial and spectral information synchronously via three directional scanning SSMs. Based on SSI and HSM, HyMamba constructs joint features from false-color and hyperspectral inputs, and enhances them through interaction with original spectral features extracted from raw hyperspectral images. Extensive experiments conducted on seven benchmark datasets demonstrate that HyMamba achieves state-of-the-art performance. For instance, it achieves 73.0\% of the AUC score and 96.3\% of the DP@20 score on the HOTC2020 dataset. The code will be released at https://github.com/lgao001/HyMamba.

\end{abstract}


%
\IEEEpeerreviewmaketitle


\section{Introduction}

\IEEEPARstart{H}{yperspectral} object tracking has emerged as a promising field in visual object tracking, owing to the rich spectral cues encoded within hyperspectral (HS) images, which reveal intrinsic material properties of the target \cite{1}. While conventional trackers based on RGB videos have achieved notable advancements in recent years, their performance deteriorates under adverse conditions such as background clutter, occlusion, and deformation \cite{3,4}. In contrast, HS data offers substantially higher spectral resolution, enabling more precise discrimination between objects of similar appearance but differing in spectral composition \cite{6}. This advantage has increasingly attracted attention to HS object tracking, especially in complex scenarios where conventional RGB trackers are prone to failure.

Most HS trackers are developed based on deep learning methods, and their performance is significantly hindered by the scarcity of large-scale, well-annotated HS video datasets. Acquiring such data is often prohibitively costly, which makes training robust models from scratch infeasible \cite{7}. To circumvent this limitation, a prevalent strategy is to leverage powerful RGB-based tracking networks pretrained on extensive RGB datasets and subsequently adapt them to the HS domain through fine-tuning \cite{8,10}. In this paradigm, HS images are typically converted into formats compatible with RGB networks, either by selecting a subset of spectral bands to form a single three-channel false-color image \cite{11,13}, or by regrouping the bands into multiple false-color images that preserve more spectral diversity \cite{14,15}. These transformed images are then fed into pretrained networks, enabling knowledge transfer from the RGB domain while partially retaining the unique spectral cues of the HS data \cite{17,18,19}. Recent advances have further explored transformer-based tracking architectures in this context, owing to their superior capacity for modeling long-range dependencies. Specifically, transformer-based RGB tracking networks are applied to extract or enhance features from band regrouping images \cite{7,8}. Superb performance is achieved by the transformer-based HS trackers due to the utilization of more spectral information.
 
Despite the progress, the development of existing HS object tracking frameworks is restricted by two crucial issues. Firstly, band selection or regrouping strategies inherently compromise spectral integrity by discarding fine-grained inter-band correlations and material-specific signatures \cite{6,18}. Specifically, the band selection strategy compresses the abundant spectral information into three-channel false-color images, leading to the loss of spectral information. Furthermore, the band regrouping strategy breaks the correlation information between bands in HS images. Secondly, current HS tracking frameworks fail to propagate the spectral features across network depths and consecutive frames. Using multi-level deep features is critical for complex scenarios, e.g., deformation and background clutter \cite{19}. However, existing methods lack cross-depth interaction information learning, and the layer-wise spectral semantics remain isolated. Moreover, the temporal coherence of spectral signatures is overlooked, as most trackers process frames independently, neglecting how spectral continuity could stabilize tracking during occlusions or fast motion. 

Recently, Mamba shows promise for sequential modeling \cite{23}, yet existing applications in object tracking focus on intra-frame refinement \cite{24} or inter-frame dynamics \cite{25}, failing to unify the cross-depth  and temporal dependencies. Specifically, for the HS object tracking task, modeling spectral information across network depths and consecutive frames with a unified Mamba-based module has not been explored. Moreover, traditional Mamba-based methods scan tokens in spatial domain and lack spectral-specific adaptation for HS object tracking task \cite{26,27}. Specifically, the scanning strategies in existing Mamba run forward or backward, which fails to synchronize spatial scanning with spectral correlation mining, leading to suboptimal spectral-spatial fusion. The resulting gap between general state space modules (SSMs) and HS demands necessitates a structured scanning strategy, which explicitly integrates forward-backward spatial modeling with spectral channel-wise processing.

To overcome these challenges, a new Mamba-based HS object tracking framework, HyMamba, is introduced, which encodes both intra-frame cross-depth spectral information and inter-frame temporal spectral information with the proposed methods. Firstly, the spectral state integration (SSI) module is designed to facilitate the joint modeling of cross-depth and temporal spectral information. SSI, which consists of a Mamba module, a joint augment (JA) module, and a spectral augment (SA) module, is deployed between transformer layers. The Mamba module is utilized to learn the semantic information from the unconverted HS image, avoiding spectral information loss in existing methods. Meanwhile, SSI iteratively updates a recurrent spectral hidden state across transformer layers and propagates temporal information across frames, achieving intra-frame cross-depth spectral information and inter-frame temporal spectral information learning. JA and SA are applied to augment the spectral information learned from the unconverted HS image and the feature extracted by the transformer layer in bidirectional manner. Secondly, hyperspectral Mamba (HSM) is proposed to further empower SSI. To learn spatial and spectral information synchronously, HSM scans the feature extracted from the unconverted HS image along forward, backward, and channel directions. Moreover, HSM augments the information learned via three-directional scanning into the features extracted by the transformer layer. Thirdly, based on SSI and HSM, HyMamba is introduced to learn cross-depth and temporal spectral information, resulting in more discriminative and temporally consistent HS tracking. HyMamba directly models cross-depth and temporal spectral information from HS images, overcoming the crucial limitations of existing HS object tracking methods.

To validate the efficacy of HyMamba, we conduct comprehensive experiments on seven HS tracking benchmarks, HOTC2020 \cite{1}, VIS2023 \cite{21}, NIR2023 \cite{21}, RedNIR2023 \cite{21}, VIS2024 \cite{22}, NIR2024 \cite{22}, and RedNIR2024 \cite{22}. These datasets span a diverse range of spectral domains and scene complexities. Across all benchmarks, HyMamba consistently achieves state-of-the-art (SOTA) performance, significantly outperforming both RGB-based and existing HS trackers in terms of accuracy and precision. The core contributions of this work are summarized as follows:

\begin{itemize} [leftmargin=1.3em]

\item We propose HyMamba, a novel HS tracking framework that explicitly models spectral, cross-depth, and temporal information in a unified manner through the SSI module and the HSM module.

\item The spectral state integration (SSI) is introduced to learn the semantic information directly from HS images, and model the cross-depth and temporal spectral information in the spectral hidden state. It avoids spectral information loss and isolation of modeling the semantic information across depth and temporal in existing HS object tracking methods.

\item The hyperspectral Mamba (HSM) is proposed to modify the general Mamba for the HS object tracking task, enabling synchronous learning of spatial and spectral information. It further enables task-specific long-range spectral dependency learning with Mamba-based module.

\item Extensive experiments on seven datasets demonstrate that HyMamba achieves superior tracking performance under various spectral settings, validating its strong generalization capability and effectiveness in HS object tracking.

\end{itemize}

\section{Related Work}

\subsection {RGB Object Tracking Methods}

Visual object tracking is an important task in computer vision (CV), and most efforts have been devoted to RGB object tracking. 
Early works formulated object tracking as a template matching task using Siamese-based network \cite{28,29}. Subsequent improvements incorporated the region proposal network \cite{30}, anchor-free \cite{32} methods, and attention mechanism \cite{35} into the Siamese-based network to advance performance. Following the development of transformer in CV, a series of methods integrated transformer modules into Siamese-based object tracking networks to model long-range dependencies in template matching \cite{40,41}.
Siamese-based methods were restricted by separated feature extractions of the template and search region. To address this, subsequent approaches jointly extracted the features of the template and search region with unified transformer-based networks \cite{3,43}.  
Building upon unified transformer-based networks, researchers introduced dynamic templates to capture the appearance variation of the target \cite{44,45}. 
Despite the progress, RGB object tracking is still restricted by the limited representative ability of RGB images in complex scenarios. Consequently, research attention has shifted toward HS object tracking, motivated by the greater representative ability of HS images.

\subsection {Hyperspectral Object Tracking Methods}

HS data enhances the tracker discriminability with abundant spectral information, leading researchers to explore various methods for effectively utilizing the spectral information \cite{1}. 
Early approaches employed traditional HS processing methods and correlation filter-based object tracking methods to achieve HS object tracking \cite{47, 52}. However, performance is unsatisfactory due to the limited semantic information learning ability. Recent HS object tracking methods applied deep learning methods for the strong semantic information learning ability, and transferred pretrained object tracking networks into the HS object tracking task to inherit the strong generalization ability \cite{12,19,53}. 
Due to the mismatch of HS images and the transferred RGB object tracking networks, \cite{17} and \cite{56} transformed the HS image into a three-channel false-color image using band selection methods. The performance remains unsatisfactory due to the loss of spectral information introduced by the band selection methods.  
Accordingly, subsequent approaches converted HS images into multiple three-channel false-color images using band regrouping methods, and processed each false-color image with a transferred object tracking network \cite{8,57}. Furthermore, \cite{11} and \cite{18} utilized feature-level fusion methods to improve the performance and efficiency of HS object tracking. 
Recently, transformer-based methods have been introduced into the HS object tracking task. Specifically, these approaches modified self-attention to enhance features and cross-attention to fuse features \cite{7,54}.
While existing methods have strong capabilities in extracting spectral features, they ignore the rich spectral cues distributed across network depths and video frames during tracking. Therefore, we incorporate Mamba into the HS tracking task, enabling both intra-frame cross-depth spectral information and inter-frame temporal spectral information learning. 

\subsection {Mamba in Visual Tracking}

Owing to its favorable trade-off between performance and efficiency, Mamba has been progressively adopted in CV \cite{26,27}. 
Several approaches have designed Mamba-based networks for feature extraction, modifying the Mamba architecture with different scanning strategies \cite{26,69,100}.
Recently, Mamba has been introduced into RGB object tracking to record the long-term temporal context information of the target in the hidden states \cite{25}. TrackMamba adopted various scanning strategies in Mamba architecture for visual object tracking \cite{66}. Mamba has also been applied to multi-modal object tracking tasks \cite{69,70}. AINet employed a Mamba-based module to fuse features from RGB and thermal images \cite{69}. 
Unlike RGB and multi-modal object tracking, the effective utilization of abundant spectral information in HS images is crucial for accurate tracking. However, the high-dimension nature of HS data restricts the application of Mamba-based methods in existing methods. To exploit the discriminative potential of spectral information, we exploit Mamba-based methods for robust HS object tracking, which models spectral information across network depth and video frames through a unified framework.

\section{Proposed Method}


\begin{figure*}
  \begin{center}
  \includegraphics[width=6.2in]{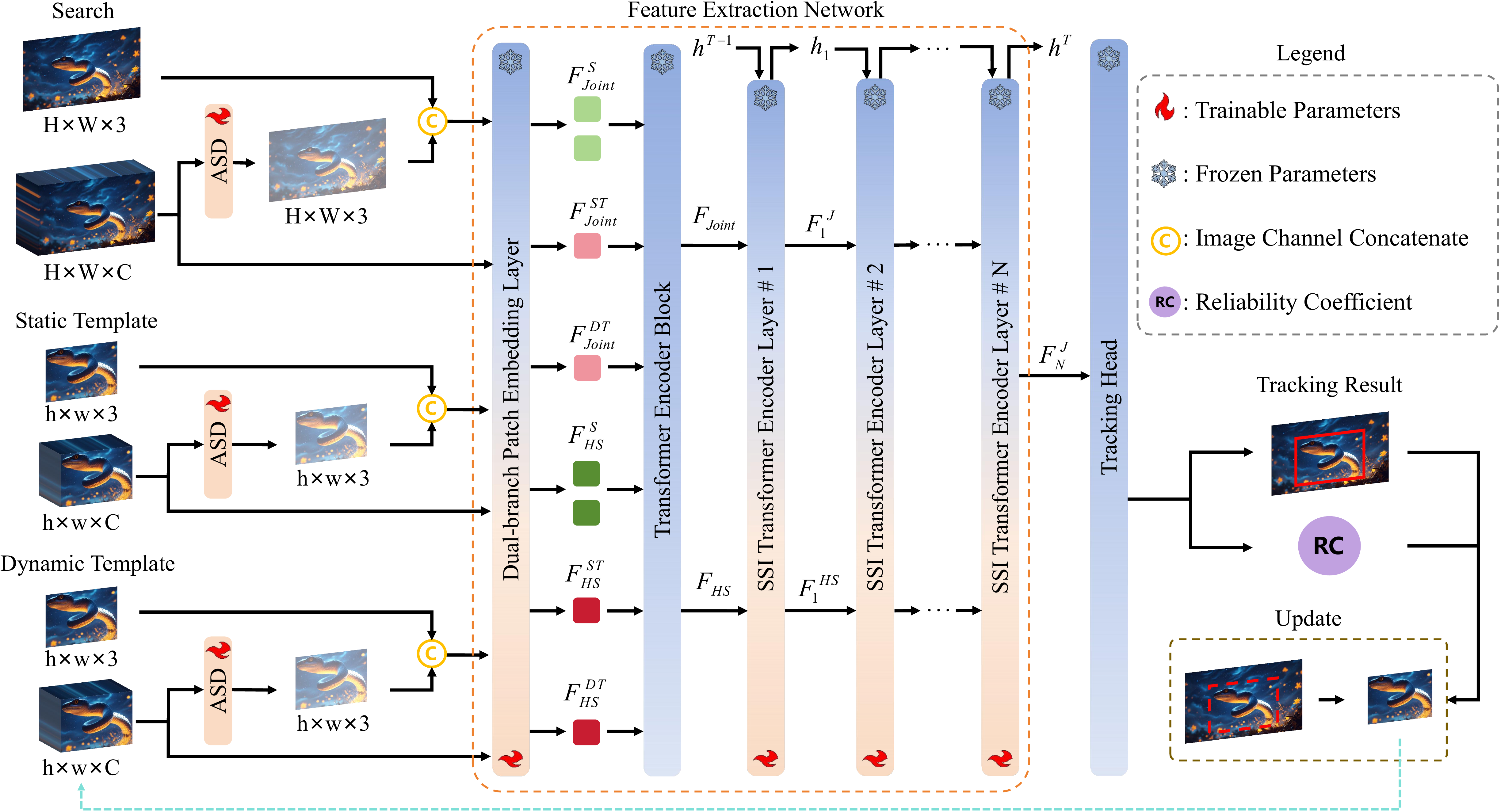}\\
 \caption{Overall architecture of the proposed HyMamba framework, comprising ASD, a feature extraction network, a tracking head, and a dynamic template update strategy.}\label{Fig1}
  \end{center}
\end{figure*}

\subsection {Overall Framework}

Fig.~\ref{Fig1} illustrates the overall architecture of HyMamba, which is composed of three primary components, an Adaptive Spectral Distillation (ASD) module for data preprocessing, a feature extraction network, and a tracking head. In the following, we elaborate on each component to elucidate the overall workflow of HyMamba.

\emph{{\textbf{1) Adaptive Spectral Distillation.}}} The proposed tracking framework takes three groups of images as input, i.e., the search group, the static template group, and the dynamic template group. The search group is cropped in current frame based on the tracking result in previous frame. The static template group preserves the target’s initial appearance from the first frame, whereas the dynamic template group is continuously updated to reflect appearance variations during tracking, thereby maintaining both identity consistency and adaptability to the changes of the target. Each group consists of a HS image and the corresponding false-color image, which is generated from the HS image with CIE \cite{1} method. Specifically, the search group contains ${X_{HS}} \in \mathbb{R} {^{H \times W \times C}}$ and ${X_{FRGB}} \in \mathbb{R} {^{H \times W \times 3}}$, while the static and dynamic template groups are denoted as ${Z_{HS}^{S}} \in \mathbb{R} {^{h \times w \times C}}$, ${Z_{FRGB}^{S}} \in \mathbb{R} {^{h \times w \times 3}}$ and ${Z_{HS}^{D}} \in \mathbb{R} {^{h \times w \times C}}$, ${Z_{FRGB}^{D}} \in \mathbb{R} {^{h \times w \times 3}}$, respectively. Here, $H \times W$ and $h \times w$ represent the height and width of the search and template images, and $C$ indicates the number of spectral channels in the HS image.

Each of the three HS inputs is first projected into a low-dimensional spectral representation using the ASD module, which is implemented with a $1 \times 1$ convolutional layer. The low-dimensional spectral representation is then concatenated with the corresponding false-color image along the channel dimension to form a unified joint representation. Taking the search group as an example, this process can be formulated as follows.
\begin{equation}\label{Eq1}
\begin{aligned}
    &\widetilde{X}_{HS} = ASD({X}_{HS}),\\
    {X}_{Joint} &= Concat(\widetilde{X}_{HS}, {X}_{FRGB}),
\end{aligned}
\end{equation}
where $\widetilde{X}_{HS} \in \mathbb{R} {^{H \times W \times 3}}$ denotes the low-dimensional spectral representation, and $Concat(\cdot )$ indicates the image channel concatenation operation. The joint representation, which integrates both spectral and spatial information, is denoted as ${X}_{Joint} \in \mathbb{R} {^{H \times W \times 6}}$. Similarly, the static template group and dynamic template group undergo the same procedure, yielding ${Z}_{Joint}^{S} \in \mathbb{R} {^{h \times w \times 6}}$ and ${Z}_{Joint}^{D} \in \mathbb{R} {^{h \times w \times 6}}$, respectively.

\emph{{\textbf{2) Feature Extraction Network.}}} As shown in Fig.~\ref{Fig1}, the feature extraction network consists of a dual-branch patch embedding layer, a pyramid transformer encoder, and a stack of SSI transformer encoder layers.

First, the dual-branch patch embedding layer consists of two parallel branches, one for the joint representations and the other for HS data. Specifically, ${X}_{Joint}$ and ${X}_{HS}$, both belonging to the search group, are partitioned into fixed-size non-overlapping patches, with each patch subsequently flattened into a 1D vector of length $D$. This process results in two feature vectors, ${F}_{Joint}^{S} \in \mathbb{R} {^{L \times D}}$ and ${F}_{HS}^{S} \in \mathbb{R} {^{L \times D}}$. The same operations are applied to the static and dynamic template groups, resulting in ${F}_{Joint}^{ST},{F}_{HS}^{ST} \in \mathbb{R} {^{l \times D}}$ and ${F}_{Joint}^{DT},{F}_{HS}^{DT} \in \mathbb{R} {^{l \times D}}$. Here, $L$ and $l$ denote the number of the search vectors and the template vectors.

In the subsequent stage, the network employs a transformer encoder block to process the features from search and template groups. This block concatenates the two sets of vectors and models them to form unified features, ${F}_{Joint} \in \mathbb{R} {^{(L+2l) \times D}}$ and ${F}_{HS} \in \mathbb{R} {^{(L+2l) \times D}}$.

Each SSI transformer encoder layer is designed to exploit the modeling capacity of transformer to facilitate feature interaction between ${F}_{Joint}$ and ${F}_{HS}$. The process of the $i$-th SSI transformer encoder layer can be formulated as follows.
\begin{equation}\label{Eq2}
\begin{aligned}
    {F}_{i}^{J}, {F}_{i}^{HS}, H_{i} = SSIEncoder({F}_{i-1}^{J}, {F}_{i-1}^{HS}, H_{i-1}), 
\end{aligned}
\end{equation}
where ${F}_{i-1}^{J} \in \mathbb{R} {^{(L+2l) \times D}}$ and ${F}_{i-1}^{HS} \in \mathbb{R} {^{(L+2l) \times D}}$ denote the joint feature and HS feature fed into the $i$-th SSI transformer encoder layer, respectively, with ${F}_{0}^{J} = {F}_{Joint}$ and ${F}_{0}^{HS} = {F}_{HS}$. $H_{i-1}\in\mathbb{R} {^{3 \times n}}$, which is progressively updated across the layers and frames, represents the spectral hidden state sending to the $i$-th layer. And, $H_{i-1}$ consists of three components, $H_{i-1}^{fwd}\in\mathbb{R} {^{n}}$, $H_{i-1}^{bwd} \in \mathbb{R} {^{n}}$, $H_{i-1}^{spec}\in\mathbb{R} {^{n}}$. These components correspond to the hidden states derived from HSM’s three directional scanning paths, i.e., forward, backward, and spectral. $n$ is the length of the hidden state maintained during each state update process. The integration of these components within $H_{i-1}$ enables the network to simultaneously preserve cross-depth and temporal spectral information. The final hidden state of the current frame is denoted as ${H}^{T}={H}_{N}$, where $N$ is the number of SSI transformer encoder layers and $N=4$ in HyMamba. The initial state ${H}_{0}$ is initialized by the final hidden state ${H}^{T-1}$ from the previous frame.

\emph{{\textbf{3) Tracking Head.}}} As the final component of the overall architecture, the tracking head receives the feature, ${F}_{N}^{J}$, from the feature extraction network, and adopts a classification and regression branch to predict the final target bounding box. Specifically, the tracking head utilizes the feature of the search region to compute a classification score map. The position with the highest classification score is selected as the coarse target location. The regression calculation estimates the width, height, and localization offsets of the target bounding box. Additionally, the classification confidence score also serves as a reliability coefficient for determining whether to update the dynamic template during inference.

\begin{figure*}
  \begin{center}
  \includegraphics[width=5.5in]{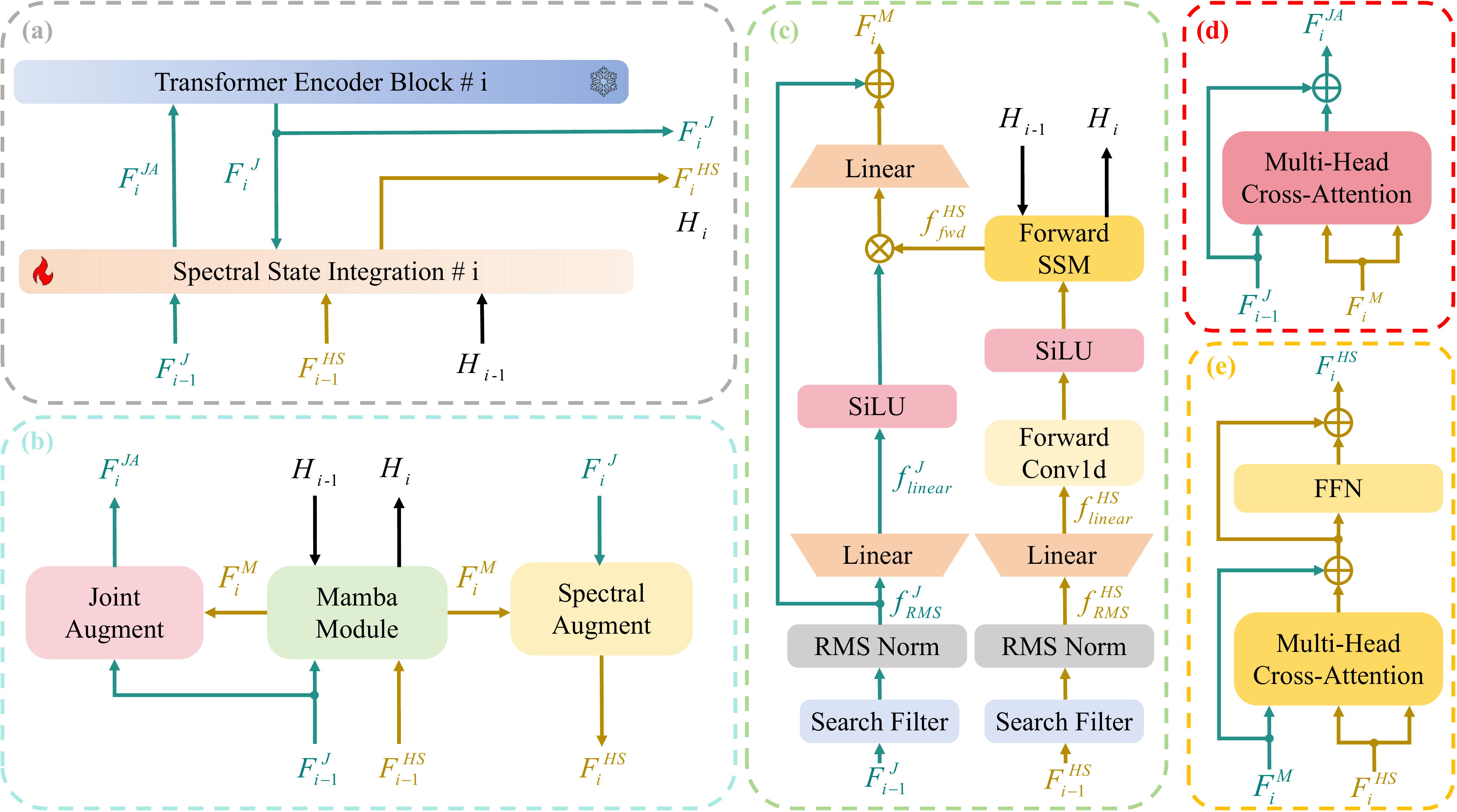}\\
 \caption{(a) Overview of a single SSI transformer encoder layer, illustrating the interaction between joint and HS features. (b) Internal composition of the proposed SSI module. (C) Structure of the mamba module. (d) Architecture of the joint augment module. (e) Design of the spectral augment module.}\label{Fig2}
  \end{center}
\end{figure*}

\subsection {Spectral State Integration Module}

HS object tacking exploits effective utilization of the spectral information for accurate tracking in complex scenarios. However, the existing methods lack cross-depth and temporal spectral information learning, restricting the performance in challenging scenarios, e.g., occlusions, fast motion and deformation.

As shown in Fig.~\ref{Fig2} (a), the $i$-th SSI transformer encoder layer integrates the bidirectional interaction between the SSI module and a transformer encoder block. The SSI module is utilized to progressively refine latent spectral-temporal states. In parallel, the transformer encoder block performs spatial and spectral feature learning jointly. In the $i$-th layer, the joint feature ${F}_{i-1}^{J}$ and the HS feature ${F}_{i-1}^{HS}$, both propagated from the previous layer, are first processed by the SSI module. Specifically, as illustrated in Fig.~\ref{Fig2} (b), SSI module consists of the Joint Augment (JA), Mamba Module (MM), Spectral Augment (SA) modules. 

The MM, which differs from standard Mamba architectures, operates on dual inputs to model spectral-temporal dependencies. Specifically, it takes both ${F}_{i-1}^{J}$ and ${F}_{i-1}^{HS}$ as inputs, leveraging their complementary information, spatial semantics from the joint feature and spectral signatures from the HS feature, to dynamically update the spectral hidden state $H_{i-1}$. As shown in Fig.~\ref{Fig2} (c), ${F}_{i-1}^{J}$ and ${F}_{i-1}^{HS}$ initially undergo the following operations.
\begin{equation}\label{Eq2.5}
\begin{aligned}
f_{RMS}^{J} =& RMS(SF(F_{i-1}^{J}),
f_{RMS}^{HS} = RMS(SF(F_{i-1}^{HS})), \\
f_{linear}^{J} &= Linear(f_{RMS}^{J}),
f_{linear}^{HS} = Linear(f_{RMS}^{HS}),
\end{aligned}
\end{equation}
where ${SF(\cdot )}$ extracts search region features, ${RMS(\cdot )}$ denotes the RMS Norm operation \cite{72}, ${Linear(\cdot )}$ denotes a dimension expanding linear projection. ${f}_{RMS}^{J} \in \mathbb{R} {^{L \times D}}$, ${f}_{RMS}^{HS} \in \mathbb{R} {^{L \times D}}$ and ${f}_{linear}^{J} \in \mathbb{R} {^{L \times 2D}}$, ${f}_{linear}^{HS} \in \mathbb{R} {^{L \times 2D}}$ represent RMS Norm-processed joint and HS features and the dimension expanded features, respectively. Then ${f}_{linear}^{HS}$ yields ${f}_{fwd}^{HS} \in \mathbb{R} {^{L \times 2D}}$ via a convolutional layer, SiLU activation \cite{73}, and forward spatial scanning in the forward SSM. After SiLU activation, ${f}_{linear}^{J}$ is weighted by ${f}_{fwd}^{HS}$, then undergoes dimension reduction via a linear layer, and is finally concatenated with ${f}_{RMS}^{J}$ through a skip connection to produce ${F}_{i}^{M} \in \mathbb{R} {^{L \times D}}$. Concurrently, the spectral hidden state is refined with $H_{i-1}$ to generate $H_{i}$. In MM, only the forward hidden state $H_{i-1}^{fwd}$ is updated. Within the JA module showing in Fig.~\ref{Fig2} (d), the joint feature, ${F}_{i-1}^{J}$, is enhanced via multi-head cross-attention (MHCA) over ${F}_{i}^{M}$, resulting in an augmented joint feature, ${F}_{i}^{JA} \in \mathbb{R} {^{(L+2l) \times D}}$. The process is formulated as follows.
\begin{equation}\label{Eq3}
\begin{aligned}
F_{i}^{JA} = F_{i-1}^{J} + MHCA(F_{i-1}^{J},F_{i}^{M},F_{i}^{M}), 
\end{aligned}
\end{equation}
where ${MHCA(\cdot )}$ denote the operations of the MHCA calculation, $F_{i-1}^{J}$ serves as the query and $F_{i}^{M}$ is used as both key and value. 

Then, $F_{i}^{JA}$ is passed through the transformer encoder block for contextual modeling, producing the refined joint feature $F_{i}^{J}$. This refined joint feature is subsequently returned to the SSI module to further enhance $F_{i}^{M}$ through the SA module. As illustrated in Fig.~\ref{Fig2} (d), this enhancement involves an MHCA calculation, in which $F_{i}^{M}$ serves as the query while $F_{i}^{J}$ functions as both key and value, and a feed-forward network. This process can be summarized as follows.
\begin{equation}\label{Eq4}
\begin{aligned}
&F_{i}^{J} = Encoder_{i}(F_{i}^{JA}), \\
F_{i}^{MHCA} &= F_{i}^{M} + MHCA(F_{i}^{M}, F_{i}^{J}, F_{i}^{J}), \\
F_{i}^{HS} &= F_{i}^{MHCA} + FFN(F_{i}^{MHCA}),
\end{aligned}
\end{equation}
where $F_{i}^{MHCA} \in \mathbb{R} {^{L \times D}}$ denotes the output of the MHCA computation within the SA module. 


\begin{figure*}[!t]
  \begin{center}
  \includegraphics[width=6in]{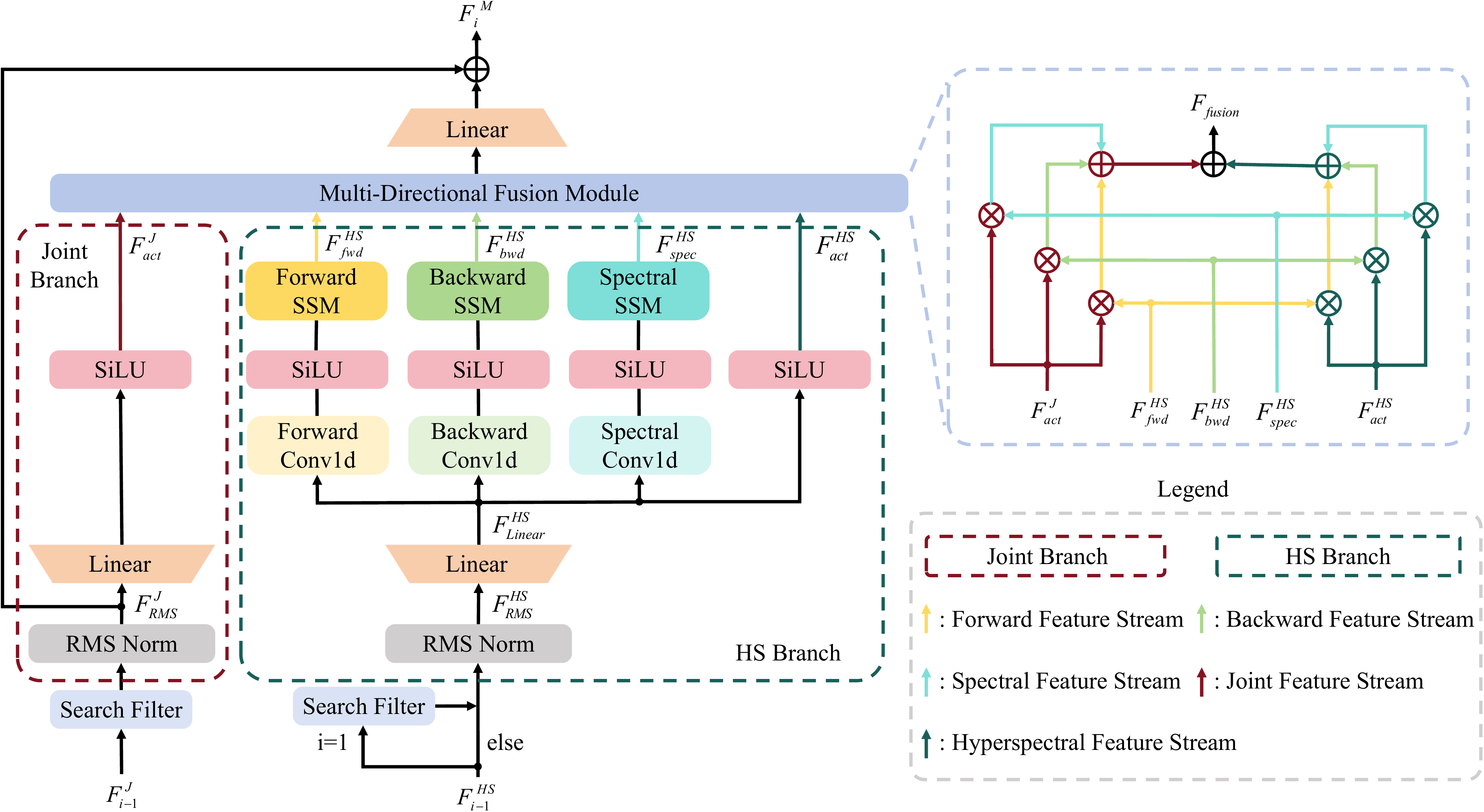}\\
 \caption{Detailed architecture of the Hyperspectral Mamba (HSM) module.}\label{Fig3}
  \end{center}
\end{figure*}

\subsection {Hyperspectral Mamba Module}

HS object tracking utilizes the spatial, spectral and temporal information in HS images to achieve accurate tracking. The MM in the SSI models the cross-depth and temporal information in the HS object tracking task with a standard Mamba module, which does not consider the spectral information modeling. To further exploit the rich spectral information embedded in HS data, the hyperspectral Mamba (HSM) is proposed to extend the scanning strategy in MM from the spatial dimension to spectral dimension, thereby facilitating the modeling of more robust and expressive spectral hidden states. As illustrated in Fig.~\ref{Fig3}, the proposed HSM integrates three distinct SSMs, i.e., a forward SSM, a backward SSM, and a spectral SSM.

Specifically, HSM takes the joint feature, ${F}_{i-1}^{J}$, and the HS feature, ${F}_{i-1}^{HS}$, from the previous layer as input. Similar to the process in MM, the two features are first processed through a search filter to obtain the search region features. Notably, for the HS feature, this step is only applied on ${F}_{0}^{HS}$. This design follows the insight that the search region, as the core of real-time localization, contains dynamic spectral variations, which are critical to tracking robustness. Subsequently, these search features are then individually normalized via RMS Norm layer, yielding ${F}_{RMS}^{J} \in \mathbb{R} {^{L \times D}}$ and ${F}_{RMS}^{HS} \in \mathbb{R} {^{L \times D}}$. Next, ${F}_{RMS}^{J}$ and ${F}_{RMS}^{HS}$ are processed in two separate branches, a joint branch and a HS branch. In the HS branch, ${F}_{RMS}^{HS}$ is first passed through a linear layer to double its channel dimension, resulting ${F}_{Linear}^{HS} \in \mathbb{R} {^{L \times 2D}}$. Meanwhile, the spectral hidden state ${H}_{i-1}$, comprising three components ${H}_{i-1}^{fwd}$, ${H}_{i-1}^{bwd}$, and ${H}_{i-1}^{spec}$, is processed together with ${F}_{Linear}^{HS}$ through three parallel SSM paths, forward SSM, backward SSM, and spectral SSM paths. Taking the spectral SSM path as an example, this process can be formulated as follows.
\begin{equation}\label{Eq5}
\begin{aligned}
F_{spec}^{HS}, &H_{i}^{spec} =  \\
&SSM_{spec}(SiLU(Conv_{spec}(F_{Linear}^{HS})),H_{i-1}^{spec}),
\end{aligned}
\end{equation}
where $F_{spec}^{HS} \in \mathbb{R} {^{L \times 2D}}$ denotes the output of the spectral path. $Conv_{spec}(\cdot )$, $SiLU(\cdot )$, and $SSM_{spec}(\cdot )$ represent the convolution computation, SiLU activation, and SSM operation in this path. The operation in SSM module can be formulated as follows.
\begin{equation}\label{Eq6}
\begin{aligned}
&{H}_{i}^{spec} = A{H}_{i-1}^{spec}+BF_{Linear}^{HS}, \\
&F_{spec}^{HS} = C{H}_{i}^{spec}+DF_{Linear}^{HS},
\end{aligned}
\end{equation}
where $A \in \mathbb{R} {^{n \times n}}$, $B \in \mathbb{R} {^{n \times 1}}$, $C \in \mathbb{R} {^{1 \times n}}$, $D \in \mathbb{R} {^{1}}$ denote the weighting parameters within the SSM, with $A$ and $B$ are discretized via the zero-order hold method \cite{23}. Following a process analogous to the spectral path, the forward and backward paths yield $F_{fwd}^{HS} \in \mathbb{R} {^{L \times 2D}}$, $H_{i}^{fwd}$ and $F_{bwd}^{HS} \in \mathbb{R} {^{L \times 2D}}$, $H_{i}^{bwd}$

\begin{figure}
  \begin{center}
  \includegraphics[width=3in]{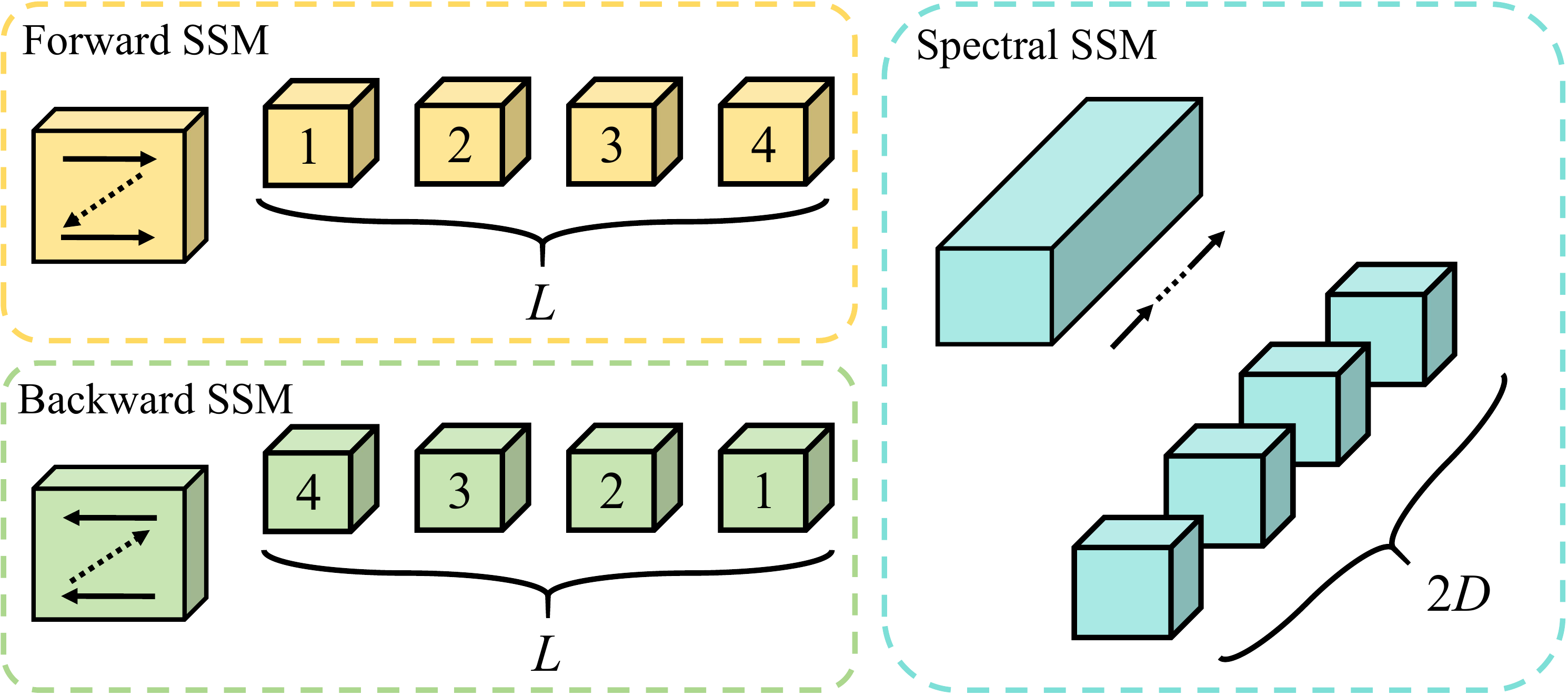}\\
 \caption{Illustration of the three scanning strategies adopted in the HSM.}\label{Fig4}
  \end{center}
\end{figure}

The scanning strategies of the three SSM are illustrated in Fig.~\ref{Fig4}. The forward and backward SSMs focus on spatial continuity, scanning along horizontal and vertical dimensions to model local and global spatial contexts, which helps preserve target shape and positional cues. In contrast, the spectral SSM is uniquely designed to scan along the channel dimension, this key design directly targets the multi-band nature of HS data. By propagating state information across spectral channels, the spectral SSM captures long-range inter-band dependencies, thereby enhancing the robustness of spectral hidden information extracted from the HS feature.

In parallel with the three SSMs, the HS branch incorporates the fourth path, in which a SiLU activation is applied to $F_{Linear}^{HS}$ and derive $F_{act}^{HS} \in \mathbb{R} {^{L \times 2D}}$. Concurrently, the joint branch processes $F_{RMS}^{J}$ through a linear layer followed by an activation function to obtain $F_{act}^{J} \in \mathbb{R} {^{L \times 2D}}$. To facilitate the integration of spatial and spectral information, the Multi-Directional Fusion Module (MDFM) is introduced to fuse $F_{act}^{J}$ and $F_{act}^{HS}$ through adaptive weighted aggregation, which can be formulated as follows.
\begin{equation}\label{Eq7}
\begin{aligned}
F_{fusion} &= F_{act}^{J}\odot F_{fwd}^{HS} + F_{act}^{J}\odot F_{bwd}^{HS} +F_{act}^{J}\odot F_{spec}^{HS} \\
+&F_{act}^{HS}\odot F_{fwd}^{HS} + F_{act}^{HS}\odot F_{bwd}^{HS} +F_{act}^{HS}\odot F_{spec}^{HS}, 
\end{aligned}
\end{equation}
where $F_{fusion} \in \mathbb{R} {^{L \times 2D}}$ is the fused feature. Finally, $F_{fusion}$ undergoes a linear projection for dimensionality reduction and is fused with $F_{RMS}^{J}$ via a skip connection, thereby injecting complementary spatial information into $F_{fusion}$ and producing the final spectral–spatial-temporal enriched feature, $F_{i}^{M}$. In parallel, $H_{i-1}$ are updated to form the new spectral hidden state $H_{i}$.

\subsection {Training and Inference}
\emph{Training.}
Given the scarcity of HS data, fully random initialization of network parameters is suboptimal. In HyMamba, only the parameters of the ASD module, SSI module, and the HS-specific patch embedding layer are trained on HS data, while the remaining parameters are initialized from the pretrained model in \cite{71} and remain frozen. The network is trained under a multi-task objective combining classification and regression. Weighted focal loss \cite{74} is adopted for classification, and L1 loss along with generalized IoU loss \cite{75} is used for bounding box regression. The total loss, $L_{total}$, is calculated as follows.
\begin{equation}\label{Eq8}
L_{total} = L_{c} + 5L_{1} + 2L_{iou}
\end{equation}
where ${L_{c}, L_{1}},$ and $L_{iou}$ correspond to classification and regression losses.

\emph{Inference.}
HyMamba uses the first frame to initialize the template, with subsequent frames serving as search regions. The update of the dynamic template is controlled by a fixed temporal interval and a confidence threshold. Moreover, to prevent erroneous target localization in the current frame from corrupting spectral hidden states and introducing noise into subsequent tracking, we further impose a confidence threshold on the hidden state update. The spectral hidden state is propagated across frames only when the classification score exceeds this threshold.

\section{Experiments}

\subsection {Experimental Setup}
\emph{{\textbf{1) Implementation Details.}}} All experiments are conducted on a server equipped with one Intel Xeon Silver 4210R CPU and two NVIDIA RTX 3090 GPUs. HyMamba is implemented in Python using the PyTorch framework. During training, only the ASD module, SSI module, and HS-specific patch embedding layers are optimized, using a learning rate of 6e-5. The entire training process spans 15 epochs, with the learning rate decayed by a factor of 10 after the 10th epoch. The AdamW optimizer is used with a weight decay of 1e-4, and the batch size is set to 14. Both the tracking head and the backbone feature extraction network adapted from HiViT \cite{77} are frozen during training. The input search images are resized to 224\(\times\)224 pixels, while the static and dynamic template images are set to 112\(\times\)112 pixels.

\emph{{\textbf{2) Dataset.}}} The proposed method is evaluated across seven HS tracking datasets, with their specifications detailed below.

\emph{HOTC Dataset.}
The dataset consists of three modalities of video data, HS, false-color, and RGB \cite{1}. The HS videos contain 16 spectral bands, while the false-color videos are generated from the HS data using the CIE color matching functions \cite{1}. The RGB videos are captured from viewpoints close to those used for the HS recordings. In total, the dataset includes 40 videos for training and 35 videos for testing. The benchmark, HOTC2020, covers 11 challenging attributes, background clutter (BC), deformation (DEF), fast motion (FM), in-plane rotation (IPR), illumination variation (IV), low resolution (LR), motion blur (MB), occlusion (OCC), out-of-plane rotation (OPR), out-of-view (OV), and scale variation (SV).

\emph{HOTC2023 Dataset.}
The HOTC2023 benchmark comprises three datasets, VIS2023 \cite{21}, NIR2023 \cite{21}, and RedNIR2023 \cite{21}, containing 16, 25, and 15 spectral channels, respectively. Each dataset includes both HS videos and their corresponding false-color counterparts. Specifically, VIS2023 consists of 55 training videos and 46 testing videos, NIR2023 includes 40 training and 30 testing videos, and RedNIR2023 provides 15 training videos and 11 testing videos.

\emph{HOTC2024 Dataset.}
The HOTC2024 benchmark introduces three HS tracking datasets, VIS2024 \cite{22}, NIR2024 \cite{22}, and RedNIR2024 \cite{22}, comprising 16, 25, and 15 spectral bands, respectively. Specifically, VIS2024 contains 111 training and 67 testing videos, NIR2024 includes 70 training and 30 testing videos, and RedNIR2024 consists of 36 training and 20 testing videos.

\begin{table}[!t] 
\centering
\setlength{\tabcolsep}{10pt} 
\begin{threeparttable}
\caption{Comparison with RGB trackers on HOTC2020 dataset}
\label{table1}
\renewcommand\arraystretch{1} 
\begin{tabular}{ccccc}
\toprule
\multirow{2}{*}{Trackers} & \multicolumn{2}{c}{RGB} & \multicolumn{2}{c}{FRGB/HS} \\
& AUC & DP@20 & AUC & DP@20 \\
\midrule
\multicolumn{5}{c}{RGB Trackers} \\
\midrule
SiamFC++\cite{80} & 0.635 & 0.865 & 0.578 & 0.820 \\
TransT\cite{40} & 0.634 & 0.897 & 0.609 & 0.854 \\
STARK\cite{41} & 0.637 & 0.900 & 0.579 & 0.814 \\
SwinTrack\cite{10} & 0.603 & 0.856 & 0.539 & 0.773 \\
SiamCAR\cite{34} & 0.590 & 0.821 & 0.554 & 0.807 \\
OSTrack\cite{42} & 0.689 & 0.913 & 0.633 & 0.864 \\
TCTrack++\cite{82} & 0.623 & 0.872 & 0.523 & 0.764 \\
MixFormer\cite{3} & 0.694 & {\color[HTML]{00B0F0}\textbf{0.926}} & 0.590 & 0.794 \\
SeqTrack\cite{43} & 0.622 & 0.906 & 0.608 & 0.861 \\
ARTrack\cite{44} & {\color[HTML]{00B0F0}\textbf{0.695}} & 0.923 & 0.660 & 0.901 \\
SMAT\cite{85} & 0.637 & 0.894 & 0.581 & 0.831 \\
AQATrack\cite{46} & 0.630 & 0.905 & 0.559 & 0.816 \\
HIPTrack\cite{45} & 0.646 & 0.903 & 0.607 & 0.865 \\
ARTrackV2\cite{86} & 0.622 & 0.912 & 0.589 & 0.878 \\
PromptVT\cite{87} & 0.679 & 0.919 & 0.601 & 0.848 \\
MCITrack\cite{25} & {\color[HTML]{FF0000}\textbf{0.701}} & {\color[HTML]{FF0000}\textbf{0.929}} & {\color[HTML]{00B0F0}\textbf{0.681}} & {\color[HTML]{00B0F0}\textbf{0.915}} \\
\midrule
\multicolumn{5}{c}{HS Trackers} \\
\midrule
MHT\cite{1} & - & - & 0.586 & 0.882 \\
BAE-Net\cite{57} & - & - & 0.606 & 0.878 \\
SSDT-Net\cite{56} & - & - & 0.639 & 0.916 \\
SEE-Net\cite{19} & - & - & 0.666 & 0.932 \\
SiamOHOT\cite{12} & - & - & 0.634 & 0.884 \\
SiamBAG\cite{13} & - & - & 0.641 & 0.904 \\
SiamHT\cite{17} & - & - & 0.621 & 0.878 \\
MMF-Net\cite{6} & - & - & 0.691 & 0.932 \\
TBR-Net\cite{90} & - & - & 0.660 & 0.920 \\
PHTrack\cite{91} & - & - & 0.660 & 0.919 \\
SPIRIT\cite{11} & - & - & 0.679 & 0.925 \\
SENSE\cite{92} & - & - & 0.690 & 0.952 \\
SpectralTrack\cite{93} & - & - & {\color[HTML]{00B0F0}\textbf{0.727}} & {\color[HTML]{00B0F0}\textbf{0.954}} \\
SP-HST\cite{94} & - & - & 0.713 & 0.952 \\
HotMoE\cite{95} & - & - & 0.704 & 0.935 \\
DaSSP-Net\cite{96} & - & - & 0.682 & 0.917 \\
\midrule
HyMamba & - & - & {\color[HTML]{FF0000}\textbf{0.730}} & {\color[HTML]{FF0000}\textbf{0.963}} \\
\bottomrule
\end{tabular}
\begin{tablenotes}
\item The red and blue colors mark the top1 and top2 results in each category.
\end{tablenotes}
\end{threeparttable}
\end{table}

\begin{figure*}[!t]
\centering
\subfloat{
    \label{Fig5_1}\includegraphics[width=1.7in]{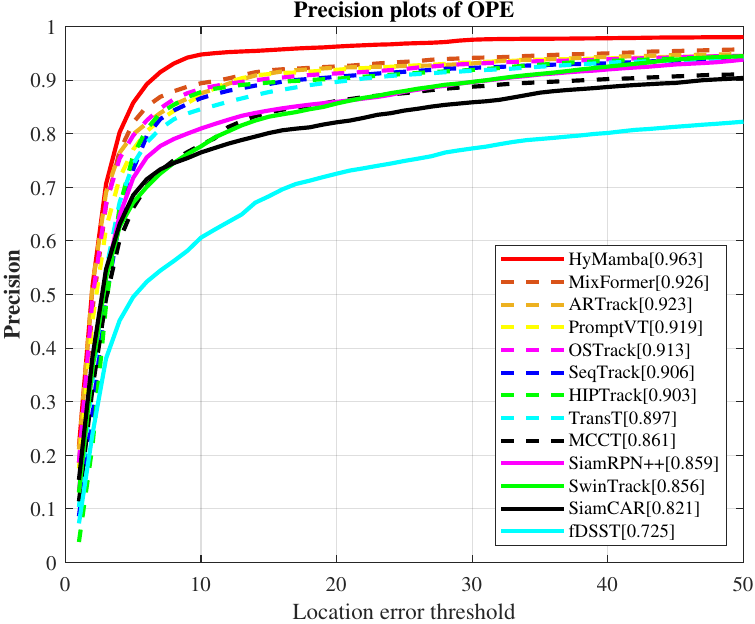}
}
\subfloat{
    \label{Fig5_2}\includegraphics[width=1.7in]{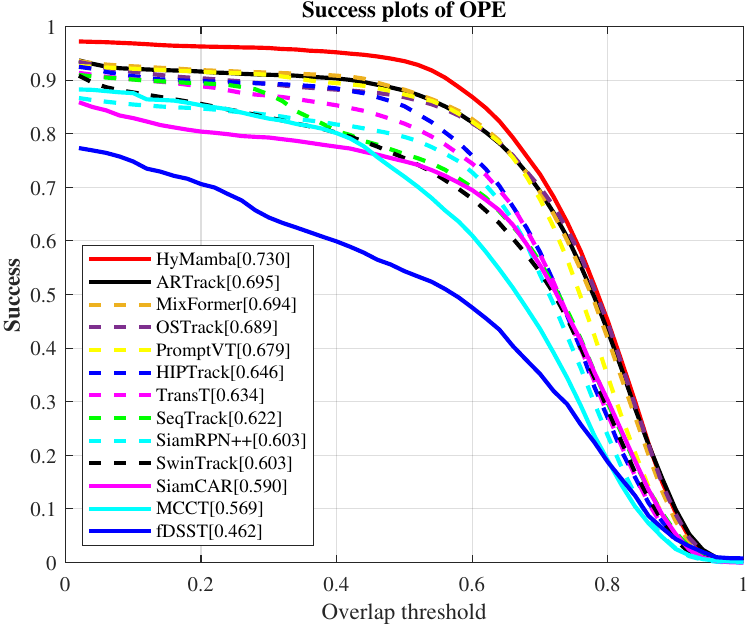}
} 
\subfloat{
    \label{Fig5_3}\includegraphics[width=1.7in]{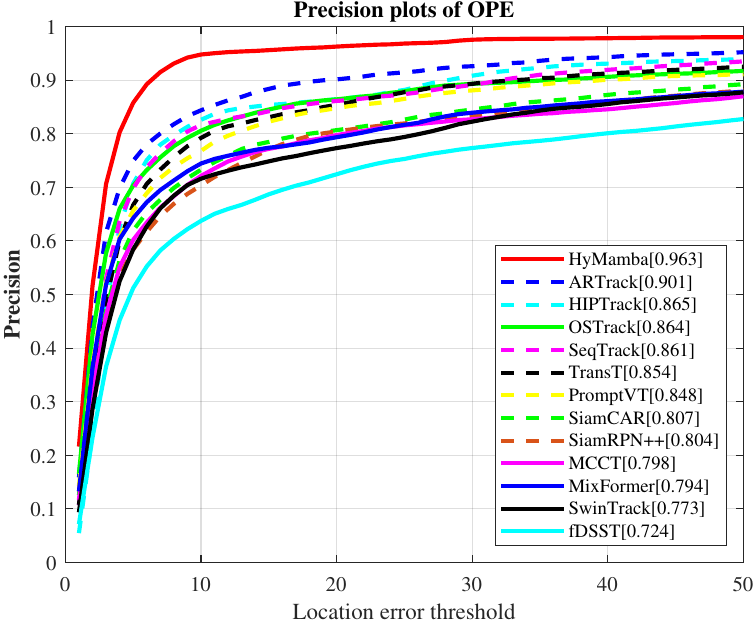}
}
\subfloat{
    \label{Fig5_4}\includegraphics[width=1.7in]{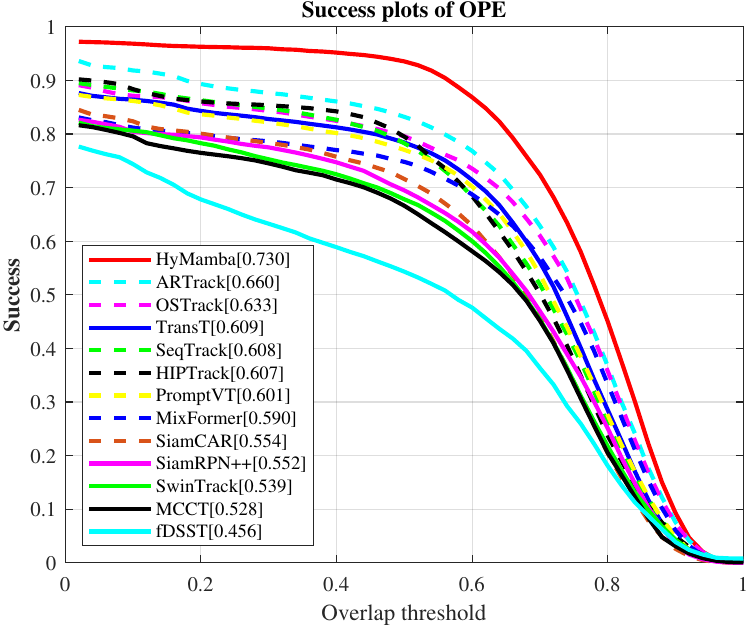}
}
\caption{Comparisons of HyMamba and RGB trackers on the corresponding RGB or false-color videos of HOTC2020 dataset.}
\label{Fig5}
\end{figure*}

\subsection {Comparison on HOTC2020 Dataset}
To validate the effectiveness of HyMamba, we conduct comparative evaluations against state-of-the-art (SOTA) RGB and HS trackers on the HOTC2020 dataset.

\emph{{\textbf{1) Comparison with RGB Trackers.}}} 
As shown in Table \ref{table1}, the performance of HyMamba is compared against 16 SOTA RGB trackers. Notably, HyMamba is evaluated on HS videos, while the RGB trackers are tested on both RGB and false-color video sequences. In Table \ref{table1}, HyMamba achieves the best overall performance, with an AUC of 0.730 and a DP@20 of 0.963. Compared to the second-best method, MCITrack \cite{25}, HyMamba surpasses it by 2.9\% on AUC and 3.4\% on DP@20 on RGB videos, and achieves 4.9\% and 4.8\% higher AUC and DP@20, respectively, on false-color videos.

Fig.~\ref{Fig5} further illustrates the precision and success plots of HyMamba and other trackers on RGB or false-color videos. As shown in Fig.~\ref{Fig5}, HyMamba ranks first and demonstrates a significant performance improvement over all RGB trackers. Compared to the most recent method, ARTrack \cite{44}, HyMamba achieves 3.5\% gain on AUC and 4.0\% improvement on DP@20. It is obvious that RGB trackers perform worse on false-color videos than on RGB videos since the transformation of HS image to false-color image results in loss of spectral information. HyMamba, which utilizes the original HS input, achieves 7.0\% improvement on AUC and 6.2\% improvement on DP@20 when compared with ARTrack on false-color videos.

\emph{\textbf{2) Comparison with HS Trackers.}} Table \ref{table1} also presents the performance comparison between HyMamba and 16 other SOTA HS trackers. HyMamba achieves the best results. Specifically, it outperforms the latest HS tracker, SP-HST \cite{94}, by 1.7\% on AUC and 1.1\% on DP@20. In addition, it achieves 0.3\% improvement on AUC and 0.9\% gain on DP@20 over SpectralTrack \cite{93}, which ranks second overall.

\subsection {Comparison on HOTC2023 Dataset}

\begin{figure*}[!t]
\centering
\subfloat{
    \label{Fig7_1}\includegraphics[width=1.1in]{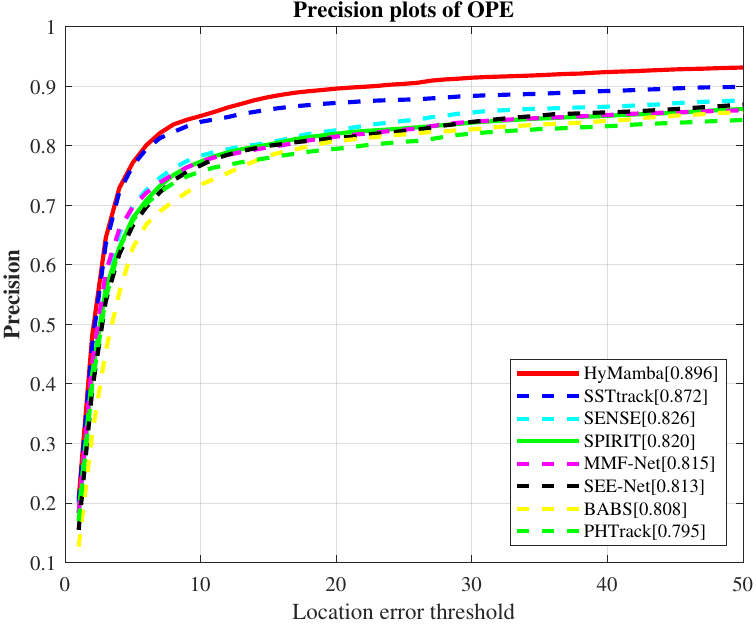}
}
\subfloat{
    \label{Fig7_2}\includegraphics[width=1.1in]{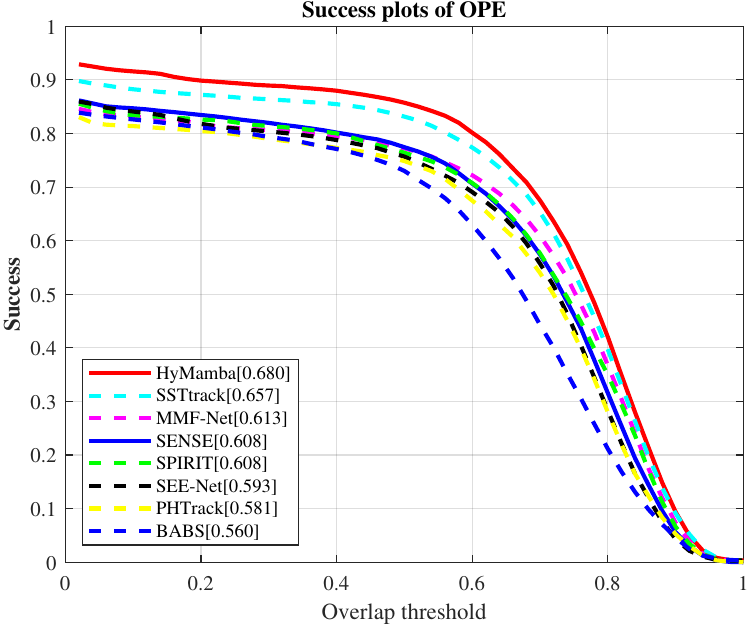}
}
\subfloat{
    \label{Fig7_3}\includegraphics[width=1.1in]{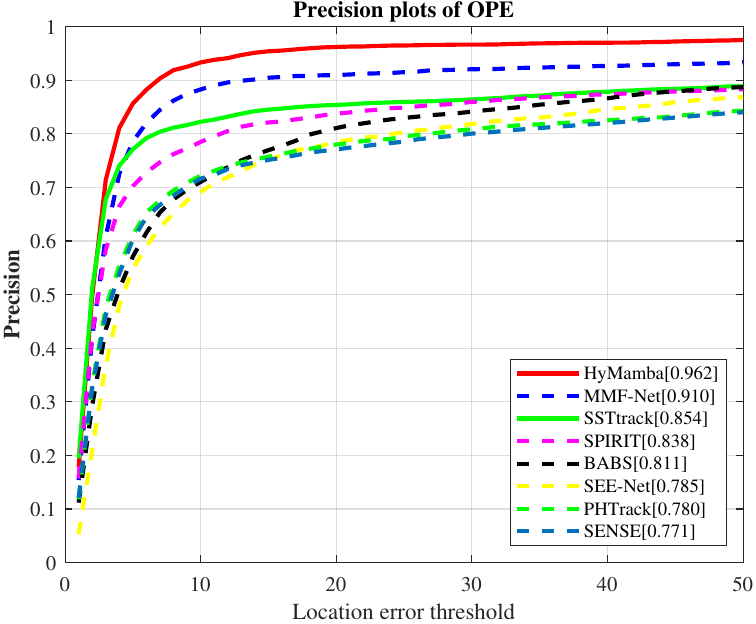}
}
\subfloat{
    \label{Fig7_4}\includegraphics[width=1.1in]{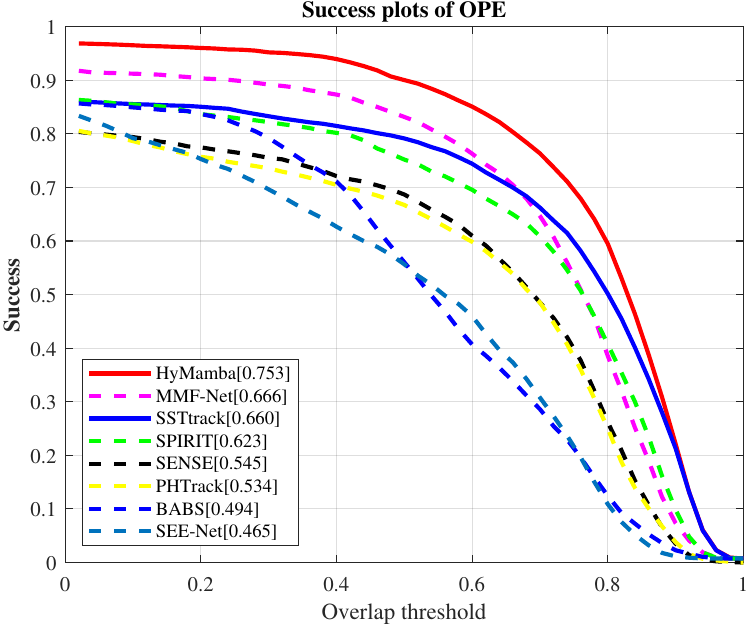}
}
\subfloat{
    \label{Fig7_5}\includegraphics[width=1.1in]{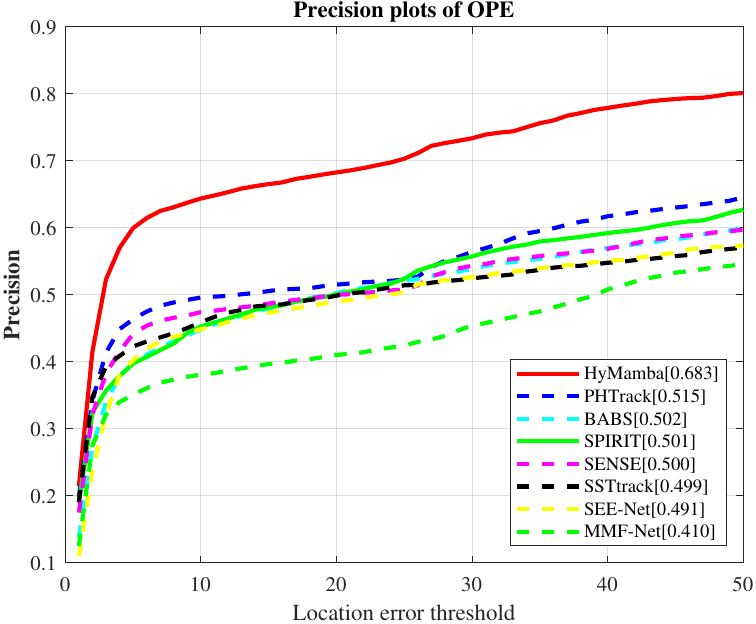}
}
\subfloat{
    \label{Fig7_6}\includegraphics[width=1.1in]{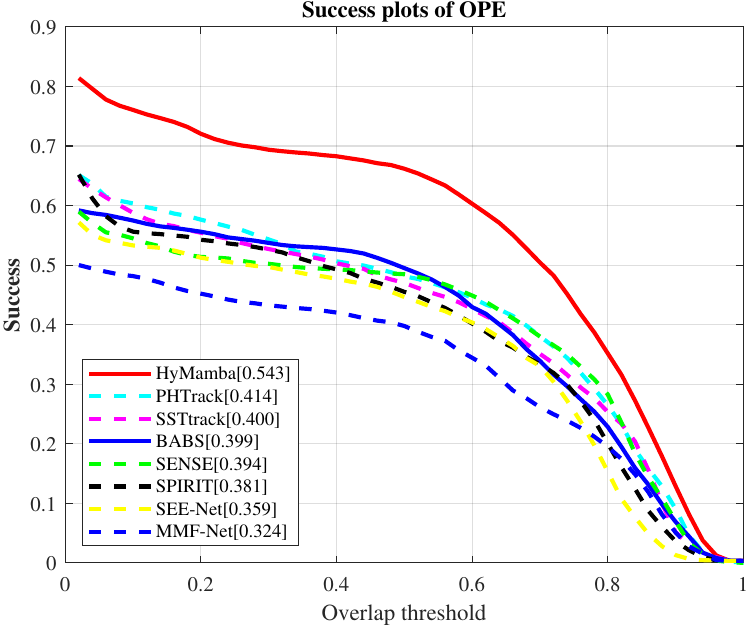}
}
\caption{Comparisons of HyMamba and other HS trackers on HOTC2023 dataset.}
\label{Fig7}
\end{figure*}

To validate the robustness of HyMamba across HS videos with varying number of spectral bands, we conduct additional experiments on VIS2023, NIR2023, and RedNIR2023. The performance of HyMamba, 8 RGB trackers and 8 HS trackers are provided in Table \ref{table2}. Fig.~\ref{Fig7} presents the precision and success plots on the three datasets.

\begin{table}[!t] 
\centering
\setlength{\tabcolsep}{4pt} 
\begin{threeparttable}
\caption{Comparison with RGB and HS trackers on HOTC2023 dataset}
\label{table2}
\renewcommand\arraystretch{1} 
\begin{tabular}{c c c c c c c} 
\toprule
\multirow{2}{*}{Trackers} & \multicolumn{2}{c}{VIS2023} & \multicolumn{2}{c}{NIR2023} & \multicolumn{2}{c}{RedNIR2023} \\
& AUC & DP@20 & AUC & DP@20 & AUC & DP@20 \\
\midrule
\multicolumn{7}{c}{RGB Trackers} \\
\midrule
SiamFC++\cite{80} & 0.536 & 0.759 & 0.637 & 0.886 & 0.371 & 0.471 \\
STARK\cite{41} & 0.533 & 0.743 & 0.431 & 0.604 & 0.378 & 0.526 \\
SiamCAR\cite{34} & 0.463 & 0.696 & 0.495 & 0.797 & 0.322 & 0.446 \\
SeqTrack\cite{43} & 0.567 & 0.809 & 0.640 & 0.872 & 0.423 & 0.554 \\
TCTrack++\cite{82} & 0.479 & 0.690 & 0.583 & 0.822 & 0.376 & 0.534 \\
SMAT\cite{85} & 0.531 & 0.744 & {\color[HTML]{00B0F0}\textbf{0.669}} & 0.866 & 0.335 & 0.448 \\
AQATrack\cite{46} & 0.537 & 0.770 & 0.599 & 0.811 & 0.445 & 0.592 \\
MCITrack\cite{25} & 0.613 & 0.829 & 0.617 & 0.814 & {\color[HTML]{00B0F0}\textbf{0.463}} & {\color[HTML]{00B0F0}\textbf{0.605}} \\
\midrule
\multicolumn{7}{c}{HS Trackers} \\
\midrule
MHT\cite{1} & 0.509 & 0.765 & 0.418 & 0.756 & 0.316 & 0.442 \\
SEE-Net\cite{19} & 0.593 & 0.813 & 0.465 & 0.785 & 0.359 & 0.491 \\
SiamBAG\cite{13} & 0.563 & 0.792 & 0.479 & 0.761 & 0.289 & 0.411 \\
MMF-Net\cite{6} & 0.613 & 0.815 & 0.666 & {\color[HTML]{00B0F0}\textbf{0.910}} & 0.324 & 0.410 \\ 
PHTrack\cite{91} & 0.581 & 0.795 & 0.534 & 0.780 & 0.414 & 0.515 \\
SPIRIT\cite{11} & 0.608 & 0.820 & 0.623 & 0.838 & 0.381 & 0.501 \\
SSTtrack\cite{97} & {\color[HTML]{00B0F0}\textbf{0.657}} & {\color[HTML]{00B0F0}\textbf{0.872}} & 0.660 & 0.854 & 0.400 & 0.499 \\
SENSE\cite{92} & 0.608 & 0.826 & 0.545 & 0.771 & 0.394 & 0.500 \\
\midrule
HyMamba & {\color[HTML]{FF0000}\textbf{0.680}} & {\color[HTML]{FF0000}\textbf{0.896}} & {\color[HTML]{FF0000}\textbf{0.753}} & {\color[HTML]{FF0000}\textbf{0.962}} & {\color[HTML]{FF0000}\textbf{0.543}} & {\color[HTML]{FF0000}\textbf{0.683}} \\
\bottomrule
\end{tabular}
\begin{tablenotes}
\item The red and blue colors mark the top1 and top2 results.
\end{tablenotes}
\end{threeparttable}
\end{table}

\emph{\textbf{1) VIS2023.}} As reported in the second column of Table \ref{table2}, HyMamba achieves the highest performance, with AUC of 0.680 and DP@20 of 0.896. Compared to the second method, SSTtrack \cite{97}, HyMamba attains 2.3\% improvement on AUC and 2.4\% gain on DP@20. As shown in Fig.~\ref{Fig7}, HyMamba consistently surpasses competing HS trackers with a noticeable margin.

\emph{\textbf{2) NIR2023.}} As shown in the third column of Table \ref{table2}, SMAT \cite{85} ranks second in terms of AUC with a score of 0.669, while HyMamba surpasses it by 8.4\% on AUC and 9.6\% on DP@20. Although MMF-Net \cite{6}, which adopts an unmixing-based strategy, pushes DP@20 to 0.910, HyMamba still outperforms it by 5.2\% on DP@20 and 8.7\% on AUC. The performance superiority of HyMamba on NIR2023 is also illustrated in Fig.~\ref{Fig7}, where it consistently dominates across the entire evaluation spectrum, highlighting its capability to extract and leverage rich spectral cues from 25-bands NIR data.

\emph{\textbf{3) RedNIR2023.}} As presented in the fourth column of Table \ref{table2}, HyMamba demonstrates the most outstanding performance, achieving an AUC of 0.543 and a DP@20 of 0.683, clearly surpassing leading trackers. As depicted in Fig.~\ref{Fig7}, HyMamba exhibits a clear advantage on the RedNIR2023 benchmark. Despite operating on only 15 spectral bands, HyMamba effectively harnesses the available spectral structure and maintains superior robustness. 

\begin{table}[!t] 
\centering
\setlength{\tabcolsep}{4pt} 
\begin{threeparttable}
\caption{Comparison with RGB and HS trackers on HOTC2024 dataset}
\label{table3}
\renewcommand\arraystretch{1} 
\begin{tabular}{c c c c c c c}
\toprule
\multirow{2}{*}{Trackers} & \multicolumn{2}{c}{VIS2024} & \multicolumn{2}{c}{NIR2024} & \multicolumn{2}{c}{RedNIR2024} \\
& AUC & DP@20 & AUC & DP@20 & AUC & DP@20 \\
\midrule
\multicolumn{7}{c}{RGB Trackers} \\
\midrule
SiamFC++\cite{80} & 0.429 & 0.615 & 0.640 & 0.864 & 0.295 & 0.403 \\
STARK\cite{41} & 0.508 & 0.665 & 0.447 & 0.590 & 0.374 & 0.499 \\
SiamCAR\cite{34} & 0.410 & 0.609 & 0.533 & 0.820 & 0.291 & 0.477 \\
SeqTrack\cite{43} & 0.525 & 0.691 & 0.632 & 0.818 & 0.370 & 0.483 \\
TCTrack++\cite{82} & 0.412 & 0.591 & 0.590 & 0.818 & 0.264 & 0.432 \\
SMAT\cite{85} & 0.460 & 0.620 & 0.680 & 0.858 & 0.323 & 0.457 \\
AQATrack\cite{46} & 0.544 & {\color[HTML]{00B0F0}\textbf{0.720}} & 0.679 & 0.856 & 0.414 & 0.543 \\
MCITrack\cite{25} & {\color[HTML]{00B0F0}\textbf{0.557}} & 0.719 & 0.688 & 0.850 & 0.441 & 0.608 \\
\midrule
\multicolumn{7}{c}{HS Trackers} \\
\midrule
MHT\cite{1} & 0.384 & 0.564 & 0.405 & 0.731 & 0.238 & 0.383 \\
SEE-Net\cite{19} & 0.400 & 0.560 & 0.509 & 0.769 & 0.387 & 0.521 \\
SiamBAG\cite{13} & 0.397 & 0.559 & 0.536 & 0.783 & 0.226 & 0.358 \\
MMF-Net\cite{6} & 0.488 & 0.645 & 0.701 & 0.875 & 0.395 & 0.521 \\
PHTrack\cite{91} & 0.306 & 0.410 & 0.526 & 0.731 & 0.264 & 0.364 \\
SPIRIT\cite{11} & 0.323 & 0.409 & 0.655 & 0.823 & 0.379 & 0.516 \\
Trans-DAT\cite{55} & 0.401 & 0.524 & 0.587 & 0.753 & 0.428 & 0.547 \\
SSTtrack\cite{97} & 0.386 & 0.482 & 0.712 & {\color[HTML]{00B0F0}\textbf{0.888}} & {\color[HTML]{00B0F0}\textbf{0.511}} & {\color[HTML]{00B0F0}\textbf{0.659}} \\
SENSE\cite{92} & 0.301 & 0.398 & 0.561 & 0.765 & 0.367 & 0.465 \\
UBSTrack\cite{98} & 0.534 & 0.681 & {\color[HTML]{00B0F0}\textbf{0.725}} & 0.861 & 0.496 & 0.627 \\
\midrule
HyMamba & {\color[HTML]{FF0000}\textbf{0.589}} & {\color[HTML]{FF0000}\textbf{0.751}} & {\color[HTML]{FF0000}\textbf{0.788}} & {\color[HTML]{FF0000}\textbf{0.939}} & {\color[HTML]{FF0000}\textbf{0.552}} & {\color[HTML]{FF0000}\textbf{0.689}} \\
\bottomrule
\end{tabular}
\begin{tablenotes}
\item The red and blue colors mark the top1 and top2 results.
\end{tablenotes}
\end{threeparttable}
\end{table}

\begin{figure*}[]
\centering
\subfloat{
    \label{Fig10_1}\includegraphics[width=1.1in]{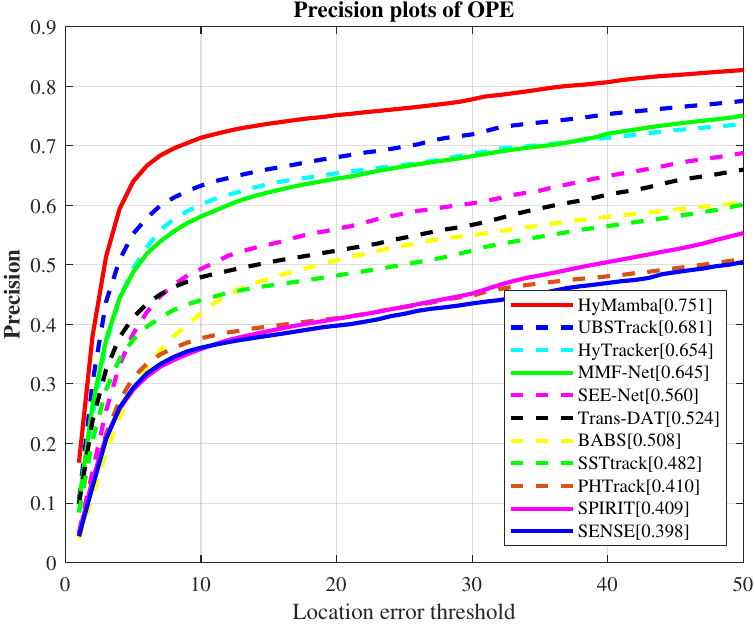}
}
\subfloat{
    \label{Fig10_2}\includegraphics[width=1.1in]{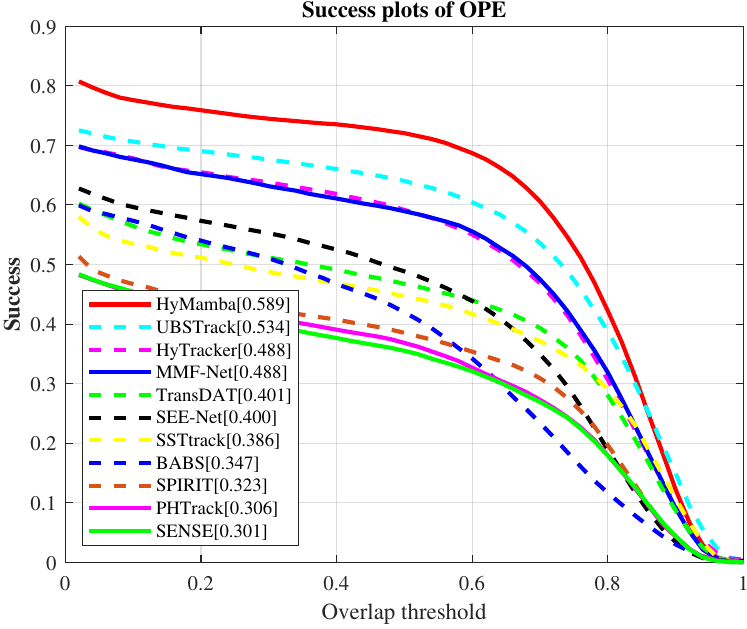}
}
\subfloat{
    \label{Fig10_3}\includegraphics[width=1.1in]{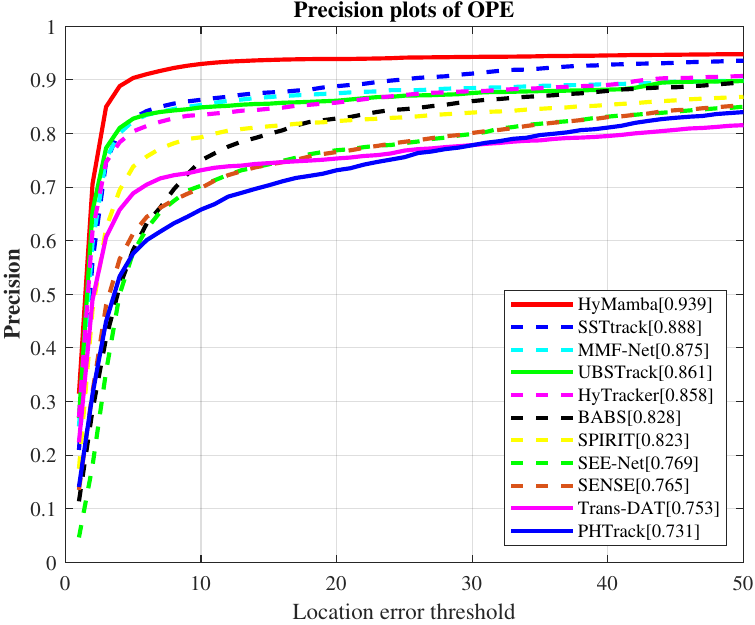}
}
\subfloat{
    \label{Fig10_4}\includegraphics[width=1.1in]{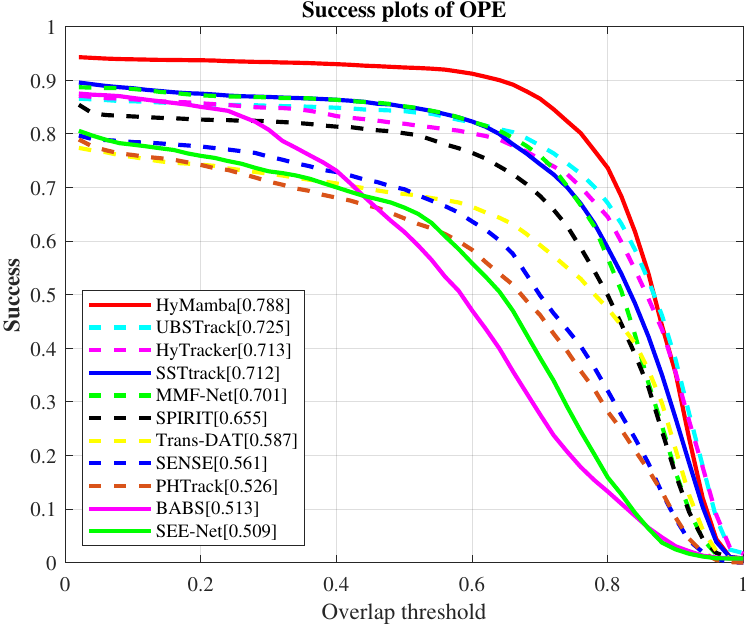}
}
\subfloat{
    \label{Fig10_5}\includegraphics[width=1.1in]{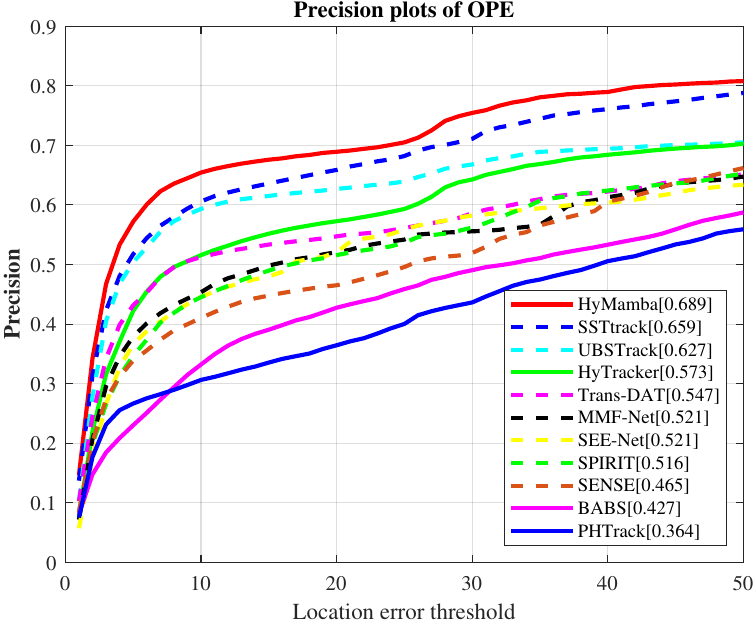}
}
\subfloat{
    \label{Fig10_6}\includegraphics[width=1.1in]{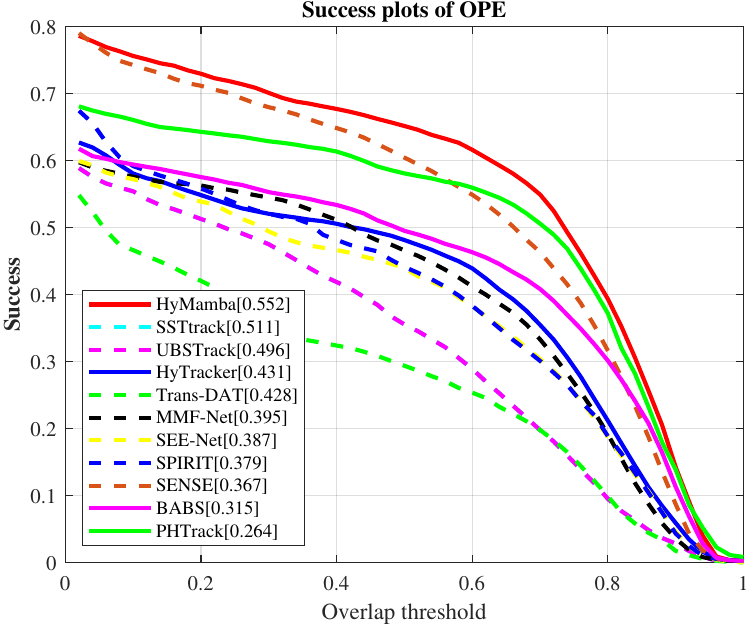}
}
\caption{Comparisons of HyMamba and other trackers on HOTC2024 dataset.}
\label{Fig10}
\end{figure*}

\subsection {Comparison on HOTC2024 Dataset}

To further verify the generalization ability and superior tracking performance of HyMamba, we conduct comprehensive evaluations on VIS2024, NIR2024, and RedNIR2024. Table \ref{table3} presents the comparative results of HyMamba against a collection of SOTA RGB and HS trackers. Fig.~\ref{Fig10} illustrates the precision and success plots on the three datasets. These results demonstrate the consistent superiority of HyMamba in tracking accuracy and robustness across various spectral conditions and scene complexities.

\emph{\textbf{1) VIS2024.}} According to the results reported in the second column of Table \ref{table3}, HyMamba achieves the top performance with AUC of 0.589 and DP@20 of 0.751. It significantly outperforms the best-performing baseline, MCITrack \cite{25}, by 3.2\% on AUC and surpasses AQATrack \cite{46}, the second-best on DP@20, by 3.1\%. As shown in Fig.~\ref{Fig10}, HyMamba consistently ranks above all competing methods across the entire overlap threshold range. 

\emph{\textbf{2) NIR2024.}} As shown in the third column of Table \ref{table3}, HyMamba again secures the best results, 0.788 on AUC and 0.939 on DP@20. In comparison, UBSTrack \cite{98}, the second-highest on AUC, lags behind by 6.3\%, while SSTtrack \cite{97}, which ranks second on DP@20, trails by 5.1\%. The quantitative plots in Fig.~\ref{Fig10} reinforce this observation, where HyMamba maintains a decisive lead throughout the evaluation curve. This performance reflects its ability to model high-dimensional spectral dependencies in the NIR domain.

\emph{\textbf{3) RedNIR2024.}} On this challenging benchmark, HyMamba continues to lead, reaching AUC of 0.552 and DP@20 of 0.689. It outperforms SSTtrack \cite{97}, the closest competitor in both metrics, by 4.1\% on AUC and 3.0\% on DP@20, demonstrating its consistent advantage across diverse spectral configurations. Fig.~\ref{Fig10} visually confirms this superiority. 

\begin{table}[!t]
\centering
\setlength{\tabcolsep}{8pt} 
\renewcommand\arraystretch{1} 
\begin{threeparttable}
\caption{Ablation Study of HyMamba on HOTC2020 Dataset}
\label{table4}
\begin{tabular}{lcccc}
\toprule
Method & AUC & $\Delta$AUC & DP@20 & $\Delta$DP@20 \\
\midrule
Baseline                 & 0.683 & -        & 0.920 & -        \\
+ ASD                   & 0.690 & +0.7\%   & 0.926 & +0.6\%   \\
+ ASD + SSI (MM)        & 0.711 & +2.8\%   & 0.934 & +1.4\%   \\
+ ASD + SSI (HSM)       & {\color[HTML]{FF0000} \textbf{0.730}} & +4.7\% & {\color[HTML]{FF0000} \textbf{0.963}} & +4.3\% \\
\bottomrule
\end{tabular}
\begin{tablenotes}
\item The red highlights indicate the best performance.
\end{tablenotes}
\end{threeparttable}
\end{table}

\subsection {Ablation Study}

To evaluate the effectiveness of the proposed method, we conduct the ablation studies on the HOTC2020 dataset.

\emph{\textbf{1) Effectiveness of individual proposed method.}} This section presents a comprehensive ablation study to assess the individual contributions of the core components within the proposed HyMamba framework. The experimental results are reported in Table \ref{table4}. The first row corresponds to a baseline based on SUTrack \cite{71}, in which only false-color inputs are concatenated and processed without any other designs. In the second configuration, the ASD module is incorporated to compress HS input into a three-channel image before concatenating with the false-color image. This strategy improves the performance by 0.7\% on AUC and 0.6\% on DP@20. Building upon this, the third row introduces the proposed SSI module, which employs MM as the hidden state extraction module. This design leads to a significant performance uplift, 2.8\% improvement on AUC and 1.4\% on DP@20, highlighting the importance of modeling the cross-depth and temporal spectral information for accurate HS object tracking. The fourth row evaluates the further contribution of HSM, achieving improvement of 4.7\% on AUC and 4.3\% on DP@20, suggesting the effectiveness of HSM.

\begin{table}[!t]
\centering
\setlength{\tabcolsep}{12pt} 
\renewcommand\arraystretch{1} 
\caption{Ablation Study of the Number of SSI on HOTC2020 Dataset}
\label{table5}
\begin{tabular}{c c c c c}
\toprule
Number & AUC & $\Delta$AUC & DP@20 & $\Delta$DP@20 \\
\midrule
0 & 0.690 & - & 0.926 & - \\
2 & 0.715 & +2.5\% & 0.947 & +2.1\% \\
4 & {\color[HTML]{FF0000} \textbf{0.730}} & +4.0\% & {\color[HTML]{FF0000} \textbf{0.963}} & +3.7\% \\
6 & 0.716 & +2.6\% & 0.939 & +1.3\% \\
8 & 0.699 & +0.9\% & 0.929 & +0.3\% \\
\bottomrule
\end{tabular}
\begin{tablenotes}
\item The red highlights indicate the best performance.
\end{tablenotes}
\end{table}

\begin{table}[!t]
\centering
\setlength{\tabcolsep}{8pt} 
\renewcommand\arraystretch{1} 
\caption{Ablation Study of the structure of SSI on HOTC2020 Dataset}
\label{table6}
\begin{tabular}{c c c c c}
\toprule
Method & AUC & $\Delta$AUC & DP@20 & $\Delta$DP@20 \\
\midrule
Without SSI & 0.690 & - & 0.926 & - \\
+ HSM & 0.697 & +0.7\% & 0.928 & +0.2\% \\
+ HSM + JA & 0.713 & +2.3\% & 0.933 & +0.7\% \\
+ HSM + JA + SA & {\color[HTML]{FF0000} \textbf{0.730}} & +4.0\% & {\color[HTML]{FF0000} \textbf{0.963}} & +3.7\% \\
\bottomrule
\end{tabular}
\begin{tablenotes}
\item The red highlights indicate the best performance.
\end{tablenotes}
\end{table}

\begin{table}[!t]
\centering
\setlength{\tabcolsep}{5pt} 
\renewcommand\arraystretch{1} 
\caption{Ablation Study of the source of spectral feature on HOTC2020 Dataset}
\label{table7}
\begin{tabular}{c c c c c}
\toprule
Method & AUC & $\Delta$AUC & DP@20 & $\Delta$DP@20 \\
\midrule
One False-color Image & 0.712 & - & 0.929 & - \\
Multiple False-color Images & 0.716 & +0.4\% & 0.940 & +1.1\% \\
Original HS Image & {\color[HTML]{FF0000} \textbf{0.730}} & +1.8\% & {\color[HTML]{FF0000} \textbf{0.963}} & +3.4\% \\
\bottomrule
\end{tabular}
\begin{tablenotes}
\item The red highlights indicate the best performance.
\end{tablenotes}
\end{table}

\begin{table}[!t]
\centering
\setlength{\tabcolsep}{4pt} 
\renewcommand\arraystretch{1} 
\caption{Ablation Study of HSM's Structure on HOTC2020 Dataset}
\label{table8}
\begin{tabular}{c c c c c}
\toprule
Method & AUC & $\Delta$AUC & DP@20 & $\Delta$DP@20 \\
\midrule
HSM & {\color[HTML]{FF0000} \textbf{0.730}} & - & {\color[HTML]{FF0000} \textbf{0.963}} & - \\
w/o Spectral SSM & 0.720 & -1.0\% & 0.946 & -1.7\% \\
w/o Spectral \& Backward SSM & 0.711 & -1.9\% & 0.934 & -2.9\% \\
w/o Joint Feature & 0.721 & -0.9\% & 0.955 & -0.8\% \\
w/o HS Feature & 0.718 & -1.2\% & 0.952 & -1.1\% \\
\bottomrule
\end{tabular}
\begin{tablenotes}
\item The red highlights indicate the best performance.
\end{tablenotes}
\end{table}

\emph{\textbf{2) Design of SSI.}} The trackers with different numbers of SSI are first evaluated, and the results are reported in Table \ref{table5}. The tracker employing 4 SSI modules yields the most favorable results. Specifically, this setup leads to an increase of 4.0\% on AUC and 3.7\% on DP@20 compared to the baseline without SSI. Reducing the number diminishes the capacity of the network to integrate cross-depth spectral information, whereas excessive stacking introduces unnecessary complexity, ultimately hampering performance. These observations validate that setting 4 SSI modules can offer optimal performance.

Further insights are provided in Table \ref{table6}, which analyzes the internal structure of the SSI module. The method in the second row integrates only the HSM. The limited increase of the performance suggests that spectral information modeling alone is insufficient to realize the full potential of HS data. Method of the third row utilizes JA, achieving 1.6\% and 0.5\% performance gains on AUC and DP@20, respectively. Extending this design further, the tracker in the fourth row incorporates SA and yields an additional performance improvement of 1.7\% on AUC and 3.0\% on DP@20. The two augmentation methods enable bidirectional enhancement of the features in the transformer encoder layer and the SSI, leading to improved performance. 

Additionally, we investigate different HS image processing strategies for SSI and report the corresponding performance in Table \ref{table7}. The first method compresses HS data into one false-color image, which discards a significant portion of spectral information. The second method partitions HS image into multiple false-color images, thereby retaining more informative content and yielding better performance than the strategy in the first row. The third strategy directly utilizes the original HS input, thereby maintaining the full spectral information and achieving the best performance.

\begin{figure}[!t]
\centering
    \label{Fig13}\includegraphics[width=3in]{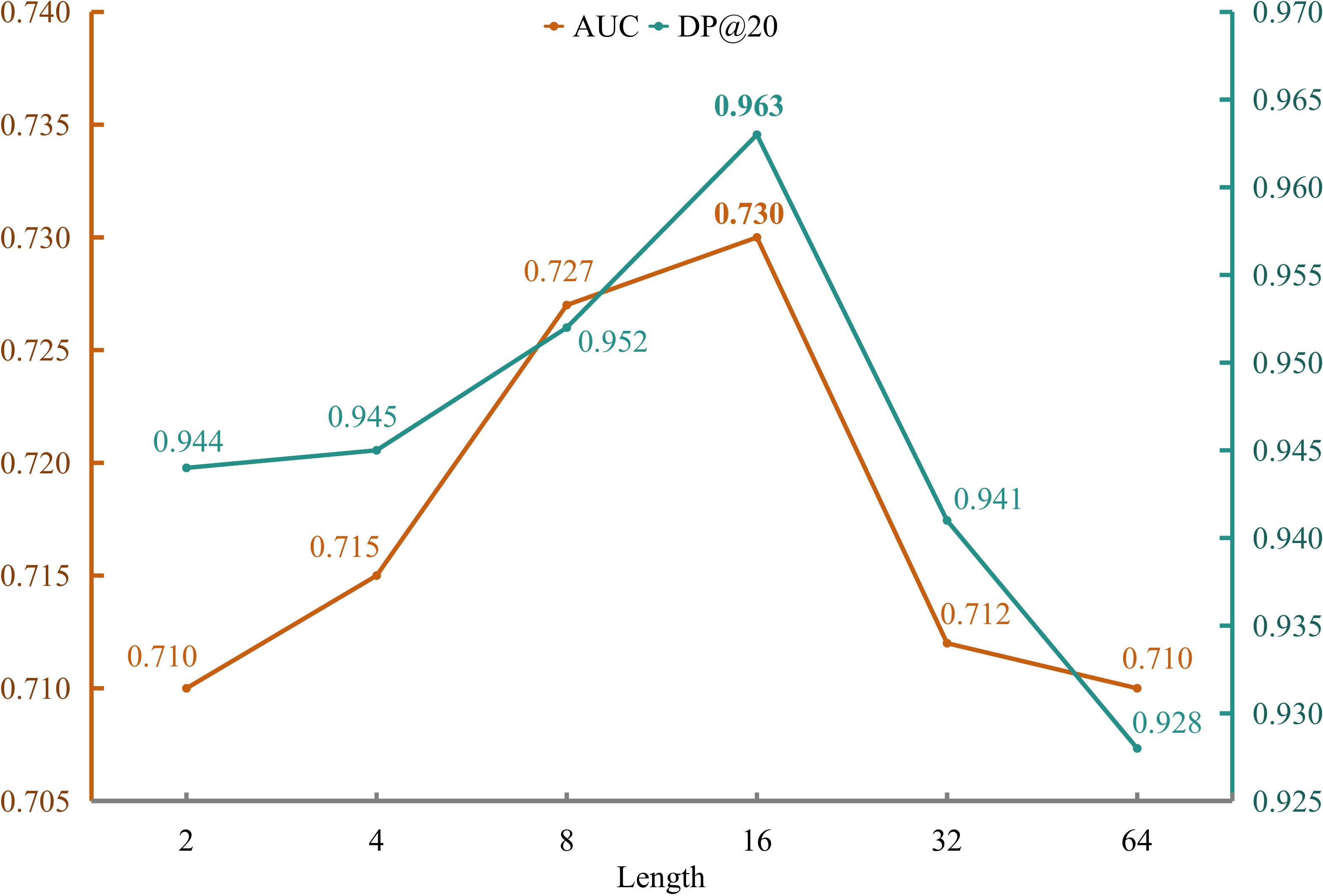}
\caption{Effect of varying spectral hidden state lengths on tracking performance.}
\label{Fig13}
\end{figure}

\emph{\textbf{3) Analysis on HSM.}} In this section, the designs of HSM are evaluated, with the corresponding results detailed in Table \ref{table8}. A comparative analysis of the first three rows unveils the critical role of the spectral SSM, whose extension of the scanning paradigm into the channel dimension allows the network to intricately capture spectral information inherent in HS data. Moreover, the performance between the first, fourth, and fifth rows highlights the superior robustness of spectral hidden state derived through cross-modal fusion. When HS features interact with joint features enriched by spatial priors, the resulting hidden states exhibit greater semantic richness and discriminative fidelity than those obtained from one modal. 

Fig.~\ref{Fig13} provides a detailed analysis of how the length of the spectral hidden state affects tracking performance. As the length increases, the AUC and DP@20 scores show a progressive improvement, culminating in peak performance at a dimension of 16. However, the performance drops when the length of the spectral hidden state is larger than 16. This can be attributed to the limited training HS data, which is inadequate to train the tracker with a long hidden state. Thus, our investigation substantiates that the length of 16 offers the best performance.

\begin{table*}[!t]
\centering
\setlength{\tabcolsep}{10pt} 
\renewcommand\arraystretch{1} 
\caption{Attributes-Based Comparison with other trackers on HOTC2020 Dataset}
\label{table9}
\begin{tabular}{cccccccccccc}
\toprule
Trackers & BC & DEF & FM & IPR & IV & LR & MB & OCC & OPR & OV & SV \\
\midrule
SiamFC++\cite{80} & 0.527 & 0.699 & 0.565 & 0.631 & 0.566 & 0.490 & 0.605 & 0.570 & 0.685 & 0.649 & 0.604 \\
SiamCAR\cite{34} & 0.484 & 0.577 & 0.526 & 0.576 & 0.333 & 0.354 & 0.575 & 0.442 & 0.613 & 0.530 & 0.498 \\
TCTrack++\cite{82} & 0.450 & 0.677 & 0.578 & 0.613 & 0.470 & 0.432 & 0.582 & 0.502 & 0.602 & 0.647 & 0.527 \\
SMAT\cite{85} & 0.548 & 0.729 & 0.632 & 0.663 & 0.515 & 0.505 & 0.688 & 0.578 & 0.713 & 0.652 & 0.597 \\
MHT\cite{1} & 0.606 & 0.664 & 0.542 & 0.670 & 0.477 & 0.475 & 0.560 & 0.564 & 0.644 & 0.626 & 0.574 \\
TSCFW\cite{52} & 0.636 & 0.648 & 0.591 & 0.724 & 0.535 & 0.548 & 0.561 & 0.556 & 0.685 & 0.654 & 0.603 \\
SiamOHOT\cite{12} & 0.699 & 0.715 & 0.560 & 0.732 & 0.517 & 0.497 & 0.610 & 0.556 & 0.728 & 0.582 & 0.627 \\
SiamBAG\cite{13} & 0.648 & 0.691 & 0.614 & 0.703 & 0.533 & 0.582 & 0.649 & 0.597 & 0.683 & 0.634 & 0.622 \\
Trans-HST\cite{59} & 0.697 & {\color[HTML]{00B0F0}\textbf{0.731}} & 0.673 & 0.729 & {\color[HTML]{00B0F0}\textbf{0.624}} & {\color[HTML]{00B0F0}\textbf{0.656}} & {\color[HTML]{00B0F0}\textbf{0.737}} & {\color[HTML]{00B0F0}\textbf{0.649}} & 0.734 & 0.648 & {\color[HTML]{00B0F0}\textbf{0.672}} \\
SiamHT\cite{17} & 0.629 & 0.682 & 0.627 & 0.702 & 0.581 & 0.576 & 0.629 & 0.580 & 0.702 & 0.650 & 0.635 \\
SPIRIT\cite{11} & {\color[HTML]{00B0F0}\textbf{0.715}} & 0.723 & {\color[HTML]{00B0F0}\textbf{0.715}} & {\color[HTML]{00B0F0}\textbf{0.760}} & 0.567 & 0.629 & 0.724 & 0.643 & {\color[HTML]{00B0F0}\textbf{0.751}} & {\color[HTML]{00B0F0}\textbf{0.689}} & 0.665 \\
HyMamba & {\color[HTML]{FF0000}\textbf{0.737}} & {\color[HTML]{FF0000}\textbf{0.781}} & {\color[HTML]{FF0000}\textbf{0.734}} & {\color[HTML]{FF0000}\textbf{0.796}} & {\color[HTML]{FF0000}\textbf{0.673}} & {\color[HTML]{FF0000}\textbf{0.683}} & {\color[HTML]{FF0000}\textbf{0.742}} & {\color[HTML]{FF0000}\textbf{0.688}} & {\color[HTML]{FF0000}\textbf{0.793}} & {\color[HTML]{FF0000}\textbf{0.776}} & {\color[HTML]{FF0000}\textbf{0.710}} \\
\bottomrule
\end{tabular}
\end{table*}

\begin{figure*}
  \begin{center}
  \includegraphics[width=5in]{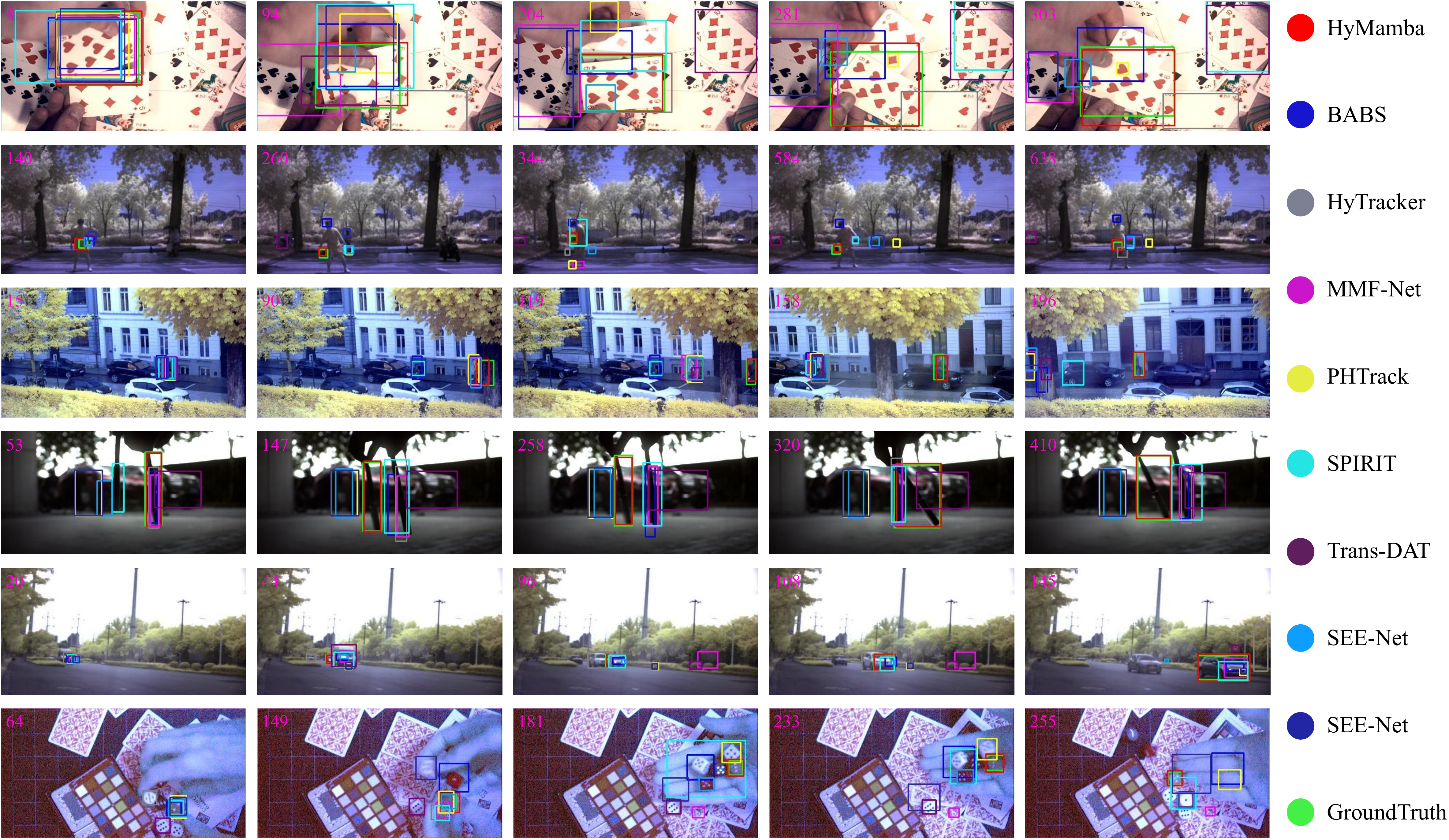}\\
 \caption{Visualization results of the proposed HyMamba and other HS trackers on six representative video sequences (vis2023-card16, nir2023-basketball3, rednir2023-rainystreet16, vis2024-pen4, nir2024-car72, and rednir2024-dice3).}\label{Fig14}
  \end{center}
\end{figure*}

\subsection {Attributes-based Evaluation}

To gain a more nuanced understanding of the tracking robustness of HyMamba across diverse real scenarios, we conduct attribute-based evaluations on the HOTC2020 benchmark, which includes 11 challenging attributes encountered in object tracking task. Table \ref{table9} summarizes the comparative results of HyMamba against 11 SOTA trackers. HyMamba achieves the highest score on all of the 11 attributes on AUC. Notably, HyMamba outperforms the second trackers by margins of 5.0\%, 4.9\%, and 8.7\% in DEF, IV, and OV, respectively. These results verify the superior adaptability of HyMamba when faced with diverse tracking challenges. 

\subsection {Visualizations of tracking results}

Fig.~\ref{Fig14} presents a visual comparison of tracking results against 8 SOTA HS trackers. The figure includes 6 videos sampled from the VIS2023, NIR2023, RedNIR2023, VIS2024, NIR2024, and RedNIR2024 datasets. In vis2023-card16, several competing methods fail to maintain the target trajectory due to the presence of visually similar distractors. Similarly, in nir2023-basketball3 and rednir2024-dice3, most trackers struggle with the diminutive scale of the targets, leading to tracking failure. In rednir2023-rainystreet16, vis2024-pen4, and nir2024-car72, the other trackers fails to track the targets due to occlusion and background clutter. However, HyMamba delivers more accurate and stable tracking results across all those sequences. 

\section{Conclusion}
In this work, we present HyMamba, a novel HS object tracking framework that unifies spectral, cross-depth, and temporal information modeling via Mamba. By introducing the SSI module, HyMamba learns the intra-frame cross-depth spectral information and inter-frame temporal spectral information from unconverted HS images with Mamba, and bidirectionally augments the learned spectral information into the features. Additionally, HSM is introduced to achieve spatial and spectral information synchronously learning with Mamba-based module. HyMamba achieves high performance on seven HS object tracking datasets, which demonstrates the effectiveness of the proposed methods and highlights the potential of the proposed methods for the accurate object tracking in complex scenarios. 

\ifCLASSOPTIONcaptionsoff
  \newpage
\fi





\bibliographystyle{IEEEtran}
\bibliography{HyA-T.bib}

\vspace{-10 mm}
\begin{IEEEbiography}[{\includegraphics[width=0.9in,clip,keepaspectratio]{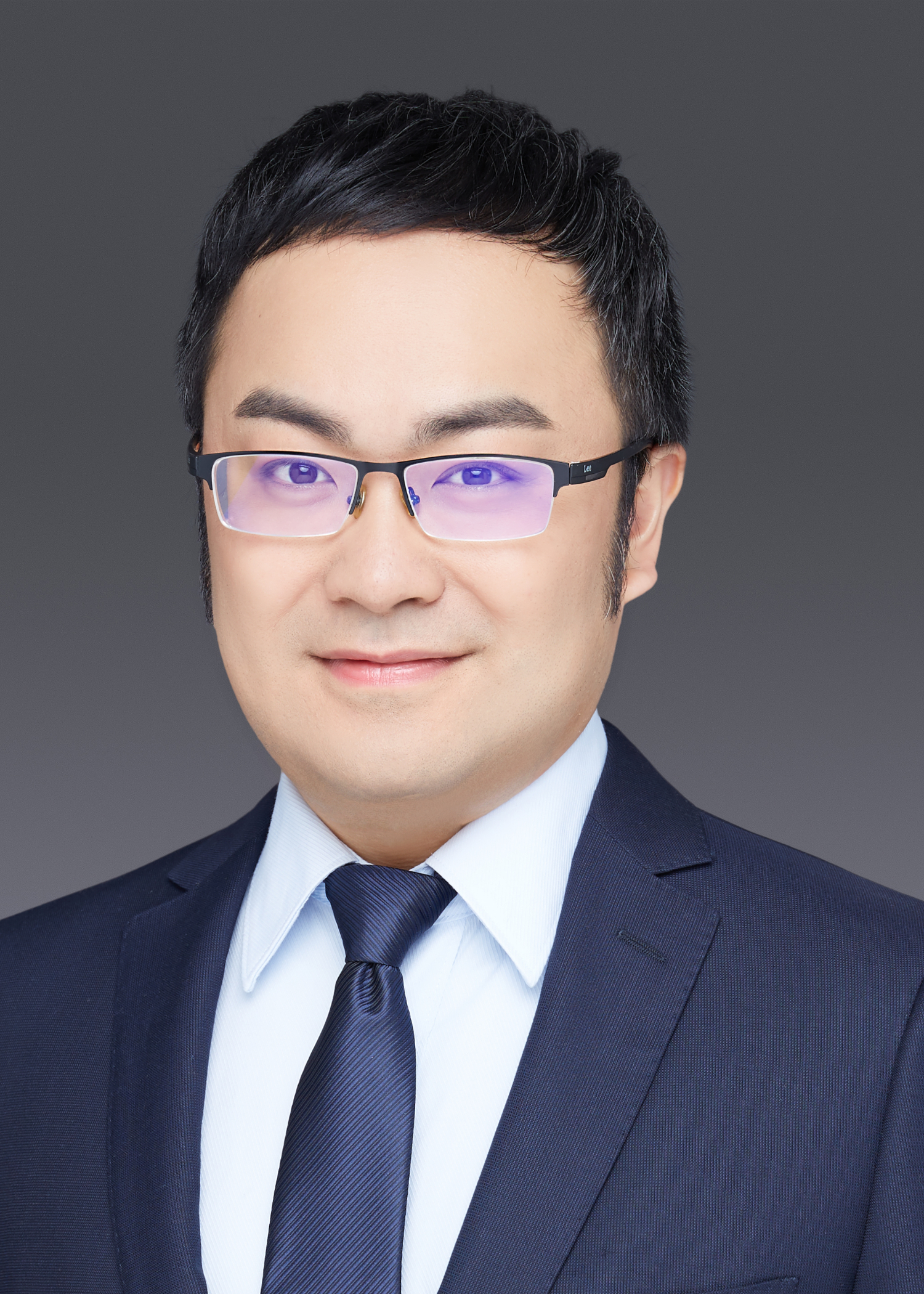}}]{Long Gao} (Member, IEEE) received the B.S. degree and the M.S. degree in Control Theory and Control Engineering from Xi'an Jiaotong University in 2010 and 2013. He received the Ph.D. degree in communication and information systems with Xidian University in 2019. Currently, he is an associate professor at communication and information systems with Xidian University. \par
His research interests include neural networks, deep learning and visual object tracking.
\end{IEEEbiography}

\vspace{-10 mm}
\begin{IEEEbiography}[{\includegraphics[width=0.9in,clip,keepaspectratio]{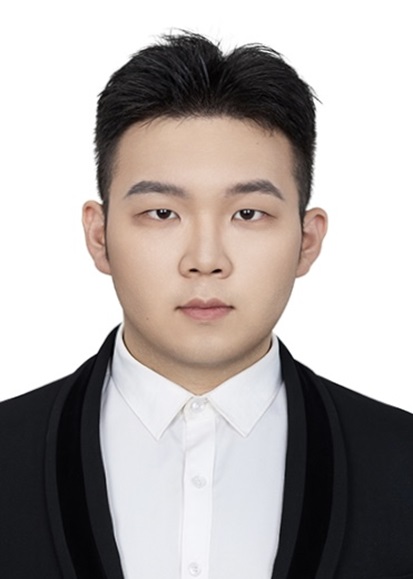}}]{Yunhe Zhang} received the B.S. degree in Communication engineering from the Tianjin University of Technology, Tianjin, China, in 2024. He is currently pursuing the M.S. degree with the Image Coding and Processing Center at State Key Laboratory of Integrated Services Networks, Xidian University, Xi’an, China. \par
His research interests include deep learning, neural network and hyperspectral object tracking.
\end{IEEEbiography}


\vspace{-10 mm}
\begin{IEEEbiography}[{\includegraphics[width=0.9in,clip,keepaspectratio]{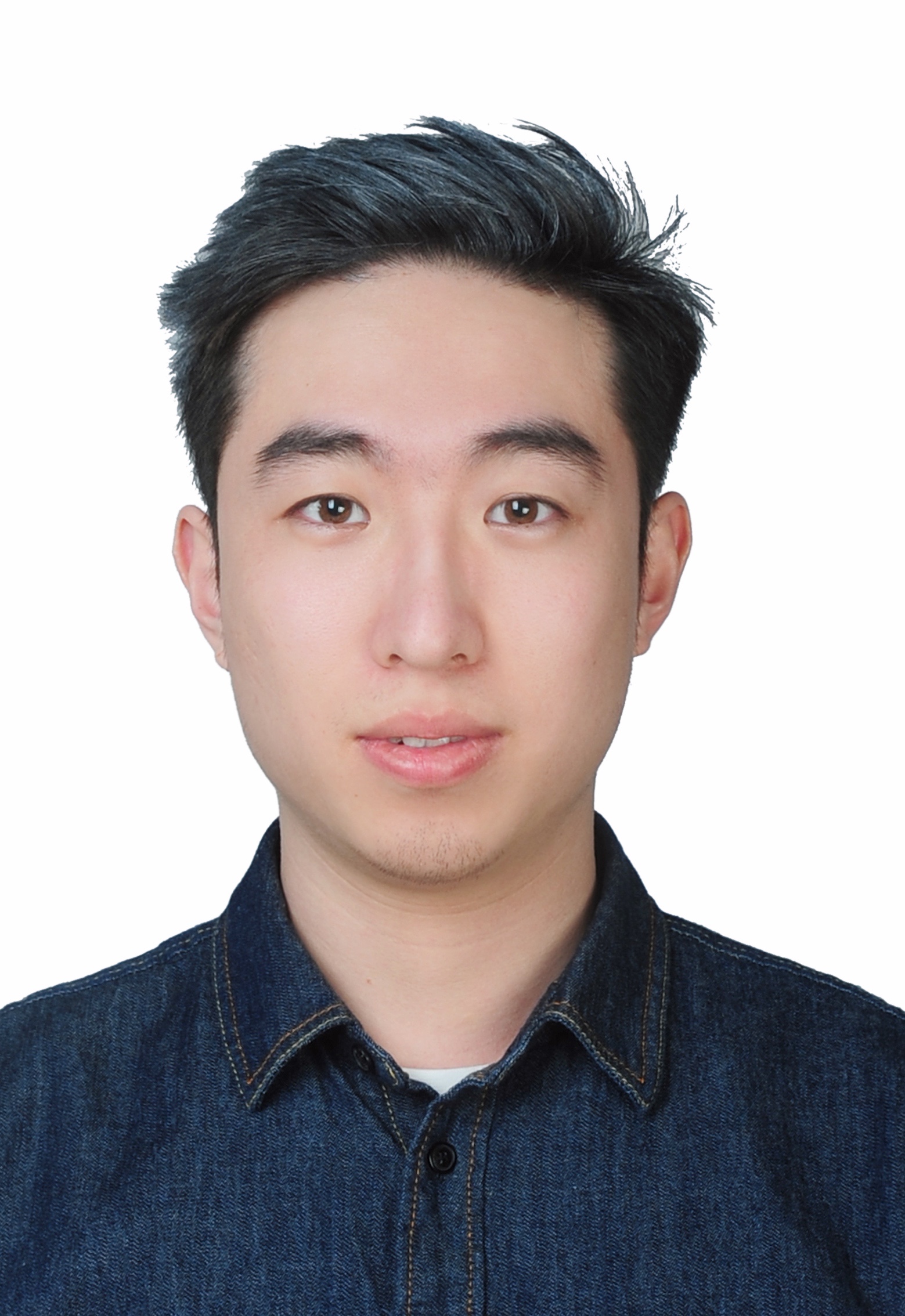}}]{Yan Jiang} received the B.E. degree in telecommunications engineering from the Xidian University, Xi’an, China, in 2016, and the M.Sc. degree in 2018 from the University of Sheffield, Sheffield, UK, where he is currently pursuing the Ph.D. degree with the Department of Electronic and Electrical Engineering. \par
His current research interests include wireless channel modeling, modulation system, mobile edge computing, smart environment modeling and deep learning.
\end{IEEEbiography}

\vspace{-10 mm}
\begin{IEEEbiography}[{\includegraphics[width=0.9in,clip,keepaspectratio]{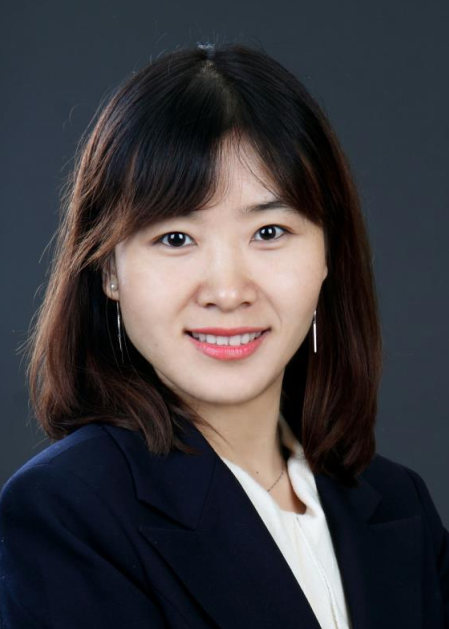}}]{Weiying Xie} (Senior Member, IEEE) received a B.S. degree in electronic information science and technology from the University of Jinan in 2011. She received an M.S. degree in communication and information systems, Lanzhou University in 2014, and the Ph.D. degree in communication and information systems of Xidian University in 2017. Currently, she is a Professor with the State Key Laboratory of Integrated Services Networks, Xidian University. \par
Her research interests include neural networks, machine learning, hyperspectral image processing, and high-performance computing.
\end{IEEEbiography}

\vspace{-10 mm}
\begin{IEEEbiography}[{\includegraphics[width=0.9in,clip,keepaspectratio]{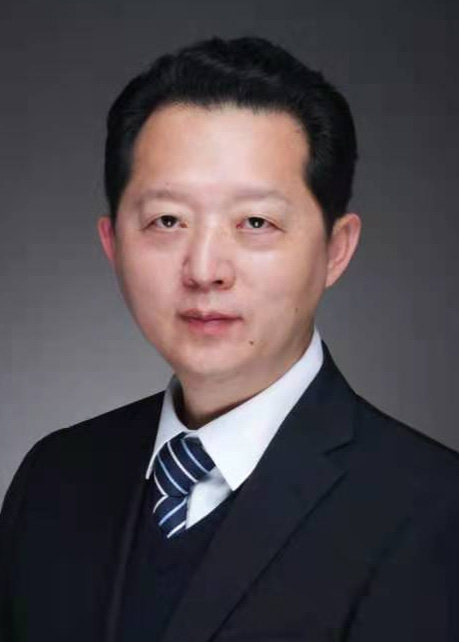}}]{Yunsong Li} (Member, IEEE) received the M.S. degree in telecommunication and information systems and the Ph.D. degree in signal and information processing from Xidian University, China, in 1999 and 2002, respectively. He joins the school of telecommunications Engineering, Xidian University in 1999 where he is currently a Professor. Prof. Li is the director of the image coding and processing center at the State Key Laboratory of Integrated Services Networks. \par
His research interests focus on image and video processing and high-performance computing.
\end{IEEEbiography}

\vfill


\end{document}